%% file: main.tex
\documentclass[11pt]{article}

\usepackage[preprint]{acl}

\usepackage{times}
\usepackage{latexsym}

\usepackage[T1]{fontenc}

\usepackage[utf8]{inputenc}

\usepackage{microtype}

\usepackage{inconsolata}

\usepackage{graphicx}

\usepackage{amsmath}
\usepackage{booktabs}
\usepackage{xcolor}
\usepackage{colortbl}

\title{SynthAVE: Scalable Synthetic Labeling for E-Commerce with LLM-Arena Validation}

\author{
    \textbf{Andrea Scarinci\textsuperscript{1}},
    \textbf{Virginia Negri\textsuperscript{1}},
    \textbf{Brayan Impata\textsuperscript{1}},
  \\
    \textbf{Suleiman Khan\textsuperscript{1}},
    \textbf{Victor Martinez\textsuperscript{1}},
    \textbf{Marcello Federico\textsuperscript{1}}
  \\
  \\
    \textsuperscript{1}Amazon
  \\
    \small{
      \texttt{\{andscar, vrgngr, biimpata, suleimkh, vicmg, marcfede\}@amazon.com}
    }
  }

\begin{document}
\maketitle

\input{sections/0_abstract}

\input{sections/1_introduction}
\input{sections/2_related_work}
\input{sections/3_dataset}
\input{sections/4_5_methods}
\input{sections/6_results}
\input{sections/8_conclusion}

\input{sections/7_limitations}

\section*{Ethics Statement}
SynthAVE is constructed from synthetic product data generated using the methodology of~\cite{negri2025attribute}, with seed products sampled from commercial e-commerce catalogs. All brand names, model identifiers, and potentially identifying product information were anonymized during generation to prevent privacy concerns and mitigate LLM biases from prior knowledge of specific products—no real customer data or personally identifiable information is present in the released dataset. Human validation was performed by domain experts specialized in e-commerce catalog systems, who established ground truth labels through careful review of disagreement cases between the synthetic generation pipeline and LLM arena predictions. SynthAVE is designed as a test set optimized for evaluation coverage: it spans 2,598 product type-attribute combinations with a minimum of 30 samples per combination, prioritizing breadth over the volume typically required for training. We encourage researchers to leverage this benchmark for downstream applications including validating catalog quality assurance systems, improving attribute coverage in product catalogs, developing robust multilingual extraction methods, and advancing LLM-as-judge evaluation methodologies. We hope SynthAVE contributes to more accurate and reliable e-commerce information systems that benefit both businesses and consumers.

\bibliography{custom}

\appendix

\input{sections/appendix_triage}
\input{sections/appendix_quality}
\input{sections/appendix_results_detail}
\input{sections/appendix_dataset_stats}
\input{sections/appendix_heatmaps}
\input{sections/appendix_prompts}
\input{sections/appendix_evaluation_settings}
\input{sections/appendix_cost}

\end{document}

%% file: sections/0_abstract.tex
\begin{abstract}
Fine-tuning large language models (LLMs) for e-commerce attribute extraction requires labeled data representative across thousands of product types, attributes, and multiple languages. This combinatorial scale translates to millions of annotations, rendering human labeling prohibitively costly. While recent work has demonstrated synthetic label generation using LLMs~\citep{negri2025attribute}, deploying such approaches at industrial scale requires integrated quality control mechanisms. We present \textbf{SynthAVE}, a large-scale benchmark for attribute-value verification spanning 12,726 products across 229 product types, 792 attributes, and 4 languages (Spanish, French, Italian, German). To validate synthetic labels at scale, we introduce a multi-LLM arena framework where each sample is evaluated by 21 judge configurations (7 model families $\times$ 3 prompts), with final labels determined via majority voting; disagreements with the synthetic label are expert-adjudicated and agreement cases are audited on a stratified sample. The majority vote ensemble agrees with expert labels at Cohen's $\kappa = 0.92$ (95.0\% agreement), while individual judges agree with one another only moderately (Fleiss' $\kappa = 0.76$)---by design, since we select for diversity. This demonstrates that diverse models with varying individual judgments aggregate into highly reliable predictions, enabling cost-effective validation at scale while concentrating expert effort on the cases where it changes the label. We estimate the resulting label quality at 97.9\%.
\end{abstract}

%% file: sections/1_introduction.tex
\section{Introduction}

Fine-tuning and evaluating large language models (LLMs) for e-commerce 
applications demands massive volumes of high-quality labeled data. 
Specifically, predicting and evaluating product attribute values from 
unstructured catalog text (e.g., product titles, descriptions, bullet 
points) requires labeled examples that maintain representativeness 
across three key dimensions: product categories (e.g., electronics, clothing, 
furniture), attributes (e.g., color, material, dimensions), and languages. 
At catalog scale—thousands of product categories, thousands of attributes, 
and multiple languages—achieving reliable model performance requires 
hundreds of labeled examples per (product category $\times$ attribute 
$\times$ language) triplet. This combinatorial requirement translates 
to millions of annotations, making human labeling prohibitively costly.

Prior work demonstrates the feasibility of using LLMs to generate 
synthetic labels for e-commerce attribute value 
extraction~\citep{negri2025attribute}. However, deploying such approaches 
at industrial scale introduces a critical challenge: how to validate 
label quality efficiently across heterogeneous attributes and languages 
without exhaustive manual review. A scalable solution must integrate 
synthetic label generation with automated quality control mechanisms.

This paper presents SynthAVE (Synthetic data for Attribute Value 
Extraction), a large-scale benchmark for attribute-value verification 
augmenting prior generation methodology~\citep{negri2025attribute} with 
scalable quality assurance. We introduce a multi-LLM 
auditing framework where each generated label is evaluated by 21 judge 
configurations (7 model families $\times$ 3 prompts), with final labels 
determined through majority voting. To mitigate systematic biases, we 
employ diverse models from different providers alongside varied prompting 
strategies. Human review is concentrated on the cases where the arena 
contradicts the synthetic label; agreement cases are auto-accepted and 
audited on a stratified sample.

Empirical evaluation demonstrates that diverse models with different biases aggregate into reliable predictions: the ensemble achieves 95.0\% agreement with expert labels (Cohen's $\kappa = 0.92$) while enabling cost-effective validation at scale.

Our main contributions are:
\begin{itemize}
    \item \textbf{SynthAVE}: A benchmark of 12,726 
    products spanning 229 product categories, 792 attributes, and 4 languages 
    for multilingual attribute-value verification.\footnote{The public release will additionally include English-language annotations produced using the same pipeline, extending coverage to 5 languages.}
    
    \item \textbf{Multi-LLM Validation Framework}: An auditing approach 
    employing majority voting across diverse judge configurations, 
    achieving 95.0\% agreement with expert evaluation while enabling 
    cost-effective validation at scale, with an analysis of panel composition.
\end{itemize}

%% file: sections/2_related_work.tex
\section{Related Work}

\label{sec:literature_review}

Information extraction (IE) from unstructured web text is a foundational capability for large-scale e-commerce systems and knowledge base construction. Common applications include product attribute extraction, entity population, catalog enrichment, and fact acquisition from heterogeneous online sources \citep{zhang2024automated, martinez2020information, dagdelen2024structured}. Recent advances in large language models (LLMs) have substantially lowered the barrier to deploying flexible extraction systems across domains and schemas \citep{zhu2024llms}. However, evaluating the quality of extracted information remains a central challenge, primarily due to the scarcity of high-quality human-annotated data and the scale at which modern IE systems operate.

Classical evaluation methodologies for IE depend on manually annotated ground truth datasets and report performance using metrics such as precision, recall, and F$_1$ score \citep{xu2024large,zhu2024llms}. While effective in controlled research settings, this paradigm does not scale to real-world web and e-commerce applications. Annotation is expensive, time-consuming, and often requires domain expertise, particularly when extraction targets complex or evolving schemas \citep{deng2024information, hsu2025leveraging}. As a result, many production pipelines operate with limited or no gold-standard evaluation data, motivating the development of evaluation methods that function under weak or missing supervision.

Existing benchmarks for product attribute extraction illustrate this tension between scale and quality. The Multi-source Attribute Value Extraction (MAVE) dataset~\citep{yang2021maveproductdatasetmultisource} provides 2.2 million attribute-value annotations across 1,257 attributes from Amazon product profiles---achieving impressive scale through automated pipelines. However, MAVE's quality assurance relies primarily on classifier-based filtering rather than comprehensive human validation of attribute-value annotations. Additionally, MAVE is limited to English and addresses extraction rather than verification tasks. These limitations motivate both the exploration of synthetic evaluation methods and the development of more rigorous validation frameworks.

In response to these constraints, recent work has explored evaluation strategies that reduce or eliminate dependence on human-labeled ground truth. \citet{seitl2024assessing} propose an automatic IE evaluation framework based on synthetic ground truth generation, in which artificially constructed structured facts are injected into natural documents and systems are evaluated on their ability to recover them. Despite their scalability, synthetic evaluation methods introduce limitations related to realism---artificially injected facts may not fully capture the ambiguity, inconsistency, and noise present in naturally occurring e-commerce text \citep{negri2025attribute}.

One response to the limits of a single LLM judge is to use a panel of them. \citet{zheng2023judging} introduced the arena paradigm, and \citet{verga2024replacing} showed that several smaller, diverse models can match one large judge. Panel size alone, however, does not guarantee proportionally more evidence: \citet{kohli2026nine} find that when judges tend to make the same mistakes, a nine-model panel can be worth little more than two genuinely independent votes. We therefore report inter-judge agreement as a measure of how far our judges' errors overlap, rather than as evidence that their votes are independent.

Another line of work investigates the use of LLMs to generate annotated datasets for downstream evaluation. By leveraging LLMs as annotators, it becomes possible to rapidly construct large quantities of labeled data where human annotation is infeasible \citep{mohta2023large,tan2024large}. In e-commerce contexts, this supports approximate evaluation of extraction pipelines across diverse product categories \citep{nadas2025synthetic}. However, automatically generated annotations may reflect model-specific biases or systematic errors, necessitating careful validation and selective human oversight \citep{de2025fine}. Pipeline design choices---decomposing extraction into multiple stages, structured output formats, and normalization rules---further improve performance \citep{jaradeh2023information, sabeh2024exploring, satyadharma-etal-2025-auto, zhang-etal-2025-leveraging-product}, but complicate evaluation when standardized benchmarks are unavailable.

%% file: sections/3_dataset.tex
\section{Task and Dataset}
\label{sec:dataset}
\noindent\textbf{Attribute-Value Verification Task.}
The attribute-value verification task requires determining whether a given attribute value can be verified from unstructured product text. Given a product's unstructured text (title, bullet points, description) and an attribute-value pair, the task is to assign one of three labels: CORRECT, INCORRECT, or UNKNOWN.

We illustrate with a concrete example. Given the product title \texttt{``Portable Laptop Stand, Adjustable Height, Aluminum Construction, Silver Finish, Compatible with 10-17 inch Devices''}, the label depends on the attribute-value pair: $(\texttt{color}, \texttt{Silver})$ is \textsc{correct} since the text states ``Silver Finish''; $(\texttt{color}, \texttt{Black})$ is \textsc{incorrect} as it contradicts the text; and $(\texttt{weight}, \texttt{1.2kg})$ is \textsc{unknown} since no weight information is present.

Training models to perform this three-way classification reliably requires labeled examples of all cases across diverse product categories, attributes, and languages. Our synthetic data generation pipeline produces such examples at scale, and our validation framework ensures their quality. The benchmark evaluates \emph{verification} of supplied pairs, not their discovery, so upstream extraction recall lies outside its scope.

\noindent\textbf{Synthetic Data Generation.}
We constructed the dataset in two stages. First, we sampled real products from a large e-commerce catalog spanning 4 languages (Spanish, French, Italian, German), ensuring categories exist in all languages with minimum 30 products per category. Second, for each sampled product, we generated synthetic variants using the controlled generation methodology of \citet{negri2025attribute}. This process creates a strict one-to-one correspondence: each synthetic product instance is paired with exactly one target attribute and one value slot. The generation pipeline produces all three label types: \textsc{correct} (value explicitly or implicitly supported by text), \textsc{incorrect} (text contradicts the value), and \textsc{unknown} (attribute cannot be determined from available text).

\begin{table}[t]
\caption{Dataset statistics by language}
\label{tab:dataset_stats}
\begin{center}
\begin{small}
\begin{sc}
\begin{tabular}{lrrrr}
\toprule
& ES & FR & IT & DE \\
\midrule
Total products & 3,159 & 3,501 & 3,007 & 3,059 \\
\% of dataset & 24.8 & 27.5 & 23.6 & 24.0 \\
Unique attributes & 636 & 675 & 609 & 673 \\
\midrule
\multicolumn{5}{l}{\textit{Label Distribution (\%)}} \\
CORRECT & 48.6 & 48.8 & 49.1 & 48.7 \\
INCORRECT & 15.3 & 15.8 & 14.5 & 14.3 \\
UNKNOWN & 36.1 & 35.4 & 36.4 & 36.9 \\
\bottomrule
\end{tabular}
\end{sc}
\end{small}
\end{center}
\vskip -0.1in
\end{table}

The label distribution in Table~\ref{tab:dataset_stats} is set by the generation protocol: attribute values are deliberately manipulated so that all three label types occur in usable proportions. Validation shifts this mix only mildly (Appendix~\ref{appendix:dataset_stats}).

The synthetic dataset comprises 12,726 products across 229 product categories, covering 792 unique attributes and 2,607 distinct product-attribute combinations. The dataset spans major product domains including electronics, apparel, home \& furniture, and sporting goods. Each category contains a minimum of 30 products, with larger categories such as NOTEBOOK\_COMPUTER reaching 209 products. Categories are evaluated on domain-specific attributes (e.g., \texttt{connectivity\_technology} for electronics, \texttt{closure.type} for apparel). Full category and attribute distributions are provided in Appendix~\ref{appendix:dataset_stats}.

\noindent\textbf{Validation Challenge.}
The synthetic pipeline reaches only 92.6\% accuracy in preliminary analysis, and metrics computed against noisy ground truth yield misleading conclusions. Yet exhaustive human review of 12,726 products across 4 languages is prohibitively expensive. Section~\ref{sec:framework} presents our solution.

%% file: sections/4_5_methods.tex
\section{Multi-LLM Validation Framework}
\label{sec:framework}

Having defined the attribute-value verification task and dataset construction process, we now address the core challenge: validating synthetic labels at scale without exhaustive human review. Building upon established generation methods~\citep{negri2025attribute}, we present a multi-LLM auditing framework that aggregates diverse model judgments into reliable quality estimates.

Our framework aggregates many LLM judgments into an estimate that approximates expert consensus. The judges are \emph{not} statistically independent---their errors are correlated, as we quantify in Section~\ref{sec:results}---so the design goal is to maximise the diversity of failure modes rather than to assume independent votes. We pursue diversity along two dimensions, and require each judge to meet qualification criteria analogous to those applied to human auditors (Appendix~\ref{appendix:qualification}).

\textbf{Judge Selection and Configuration.} We diversify judges along two axes.

\textit{Model family diversity.} We selected 7 LLM families from different providers (Table~\ref{tab:model_details}): Claude~\citep{TheC3}, Nova~\citep{nova_models}, GPT~\citep{openai2024gpt4technicalreport}, Mistral~\citep{mistral_large3_2025}, DeepSeek~\citep{Guo_2025}, Qwen~\citep{bai2025qwen3vltechnicalreport}, and Gemma~\citep{gemmateam2025gemma3technicalreport}. This ensures diversity across model scales and training objectives (general-purpose, reasoning-focused, multilingual), reducing the influence of any single family's biases on the aggregate.

\begin{table}[t]
\caption{LLM models evaluated in our arena.}
\label{tab:model_details}
\begin{center}
\begin{small}
\begin{sc}
\begin{tabular}{ll}
\toprule
Model Family & Model Name \\
\midrule
Anthropic & Claude 3.5 Sonnet v2 \\
Amazon & Nova Pro v1 \\
OpenAI & GPT OSS 120B \\
Mistral & Mistral Large 3 \\
DeepSeek & DeepSeek R1 \\
Qwen & Qwen3 VL 235B \\
Google & Gemma 3 27B \\
\bottomrule
\end{tabular}
\end{sc}
\end{small}
\end{center}
\vskip -0.1in
\end{table}

\textit{Prompt diversity.} For each model, we developed three substantially different prompt versions (detailed in Appendix~\ref{appendix:prompts}). This prevents model outputs from reflecting prompt over-fitting or artificial consistency due to identical phrasing.

This design yields $7 \times 3 = 21$ judge configurations. Each configuration evaluates all 12,726 products across 4 languages (Spanish, French, Italian, German), producing 267,246 total judgments.

\begin{figure}[t]
\centering
\includegraphics[width=\columnwidth]{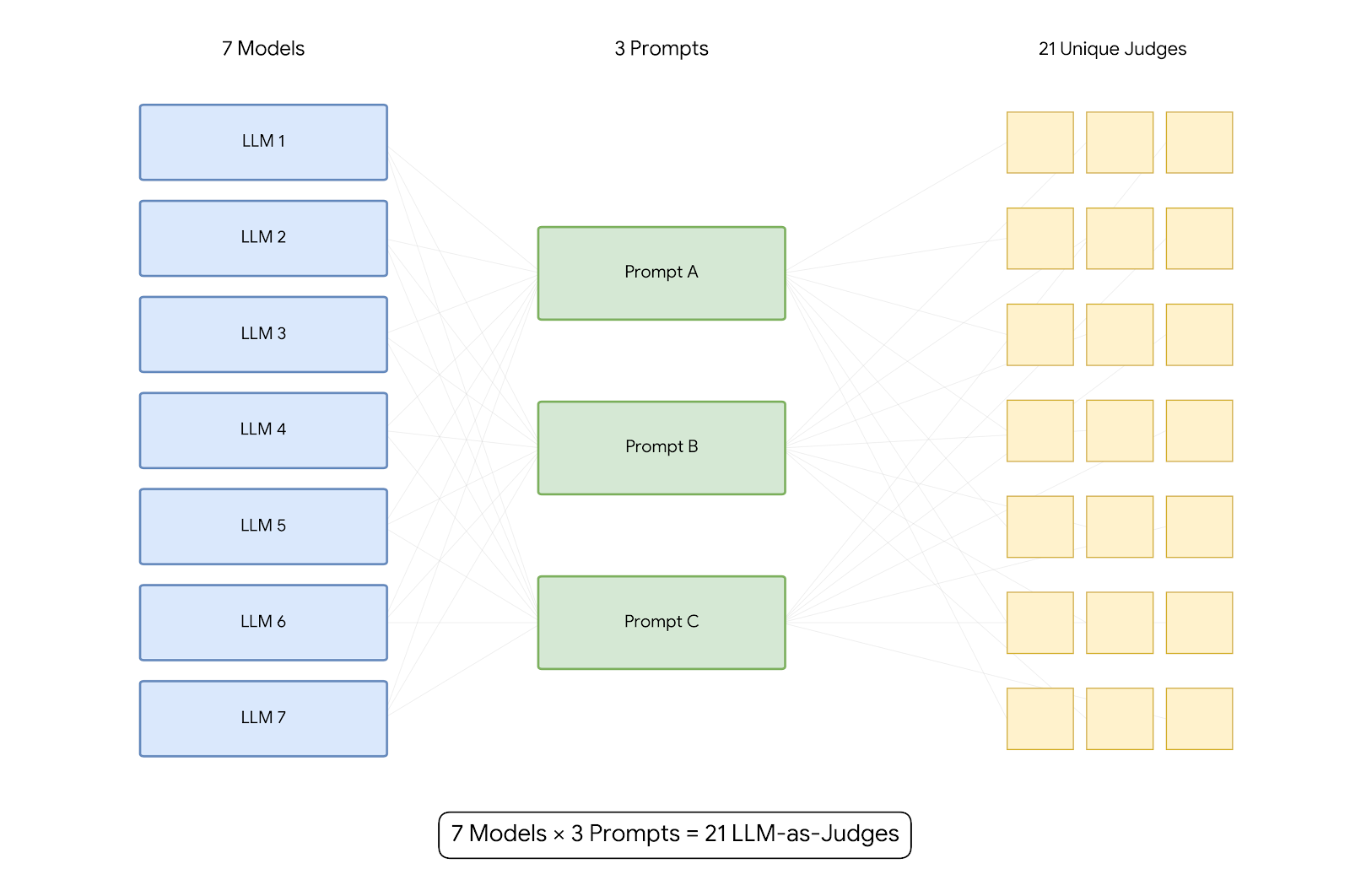}
\caption{Framework for multi-LLM judge configurations. Seven model families combined with three prompt variations yield 21 judge configurations.}
\label{fig:llm_arena}
\end{figure}

\textbf{Evaluation Task.}
Judges perform the attribute consistency 
verification task defined in Section~\ref{sec:dataset}, classifying each sample as CORRECT, INCORRECT, or UNKNOWN with brief justification.

\textbf{Aggregation through Majority Voting.}
We aggregate judge outputs via majority voting across all 21 configurations, resolving the rare ties as described in Appendix~\ref{appendix:eval_setting}. The agreement level is itself informative: unanimous votes mark the highest-confidence predictions, while split decisions flag samples warranting inspection. The signature we seek is moderate inter-judge agreement (Fleiss' $\kappa$) combined with high ensemble--expert agreement (Cohen's $\kappa$), since this indicates judges with differing failure modes cancelling errors rather than duplicating one another. Fleiss' $\kappa$ is therefore reported throughout as a diversity diagnostic, not as evidence of statistical independence.

\textbf{Ground Truth Establishment.}
To evaluate our framework, we established ground truth labels through a disagreement-based annotation strategy that leverages the LLM arena to minimize human effort. Each synthetic sample has an original label assigned by the generation algorithm. We compared this label against the LLM arena majority vote and applied the following protocol:

\begin{itemize}
    \item \textbf{Agreement cases} (11,310; 88.9\%): the majority vote matched the original synthetic label, and the sample was accepted \emph{without} human review.
    \item \textbf{Disagreement cases} (1,416; 11.1\%): the majority vote contradicted the original label, and the sample was adjudicated by a domain expert who assigned the final label.
\end{itemize}

Only the disagreement cases are therefore directly adjudicated; 10,910 samples (86\%) are never inspected by a human, so we describe cases as \textit{expert-adjudicated}, \textit{automatically accepted}, or \textit{independently audited} rather than uniformly human-validated.

To validate the agreement assumption, an expert annotator independently triaged a stratified random sample of 400 agreement cases (100 per language, comprising 100 unanimous and 300 mixed-agreement samples). The annotator overturned the majority vote in 12/400 cases (3.0\%, 95\% CI [1.3\%, 4.7\%]); six further cases were marked \textit{unsure} and counted as non-overturns, giving a conservative upper bound of 4.5\%. Unanimous cases---where all 21 judges agreed---were never overturned (0/100), and all 12 overturns occurred in mixed-agreement samples (4.0\% overturn rate). Among majority labels, UNKNOWN showed the highest overturn rate (5.0\%) compared to CORRECT (2.2\%) and INCORRECT (0.0\%), consistent with UNKNOWN being the ``hedge'' vote that occasionally masks determinable cases. Agreement between the generation algorithm and a diverse judge panel is therefore a reliable, though not infallible, signal of label correctness (full analysis in Appendix~\ref{appendix:triage}).

%% file: sections/6_results.tex
\section{Results}
\label{sec:results}

We first establish that our validation framework approaches expert-level reliability, then present the final dataset features after cleaning.

\paragraph{Overall Arena Performance.}
Table~\ref{tab:arena_performance} summarizes agreement between LLM Arena majority voting and expert evaluation. The arena achieves 95.0\% overall agreement with experts, with Cohen's $\kappa = 0.92$ indicating almost perfect inter-rater reliability~\citep{landis1977measurement}. Performance is consistent across languages (94.0--96.4\%). Individual judges agree with one another only substantially (Fleiss' $\kappa$ = 0.757), yet the majority vote ensemble reaches almost perfect agreement with experts---the intended outcome of selecting judges for diversity rather than redundancy. This figure is not fully independent of the arena, since agreement cases are accepted because the arena concurred with the synthetic label; we bound the resulting optimism below.

\begin{table}[t]
\caption{LLM Arena agreement with expert evaluation by language}
\label{tab:arena_performance}
\begin{center}
\begin{small}
\begin{sc}
\setlength{\tabcolsep}{3.5pt}
\begin{tabular}{lrrrr}
\toprule
Lang. & Products & Acc. & Cohen $\kappa$ & Fleiss $\kappa$ \\
\midrule
IT & 3,007 & 96.4\% & 0.94 & 0.755 \\
DE & 3,059 & 95.3\% & 0.92 & 0.757 \\
ES & 3,159 & 94.7\% & 0.91 & 0.757 \\
FR & 3,501 & 94.0\% & 0.90 & 0.759 \\
\midrule
\textbf{All} & \textbf{12,726} & \textbf{95.0\%} & \textbf{0.92} & \textbf{0.757} \\
\bottomrule
\end{tabular}
\end{sc}
\end{small}
\end{center}
\vskip -0.1in
\end{table}

\paragraph{Data Cleaning Performance.}
Table~\ref{tab:cleaning_performance} summarizes the Arena's effectiveness as a data cleaning mechanism. The original synthetic pipeline achieves 92.6\% accuracy (946 errors across 12,726 samples). The arena corrects 786 of these errors (83.1\% error correction rate), improving accuracy to 95.0\%. When framed as binary bad-label detection, the Arena achieves 98.0\% precision and 95.2\% recall (F1 = 96.6\%)---when it flags a label as incorrect, it is almost always right.

\begin{table}[t]
\caption{LLM Arena data cleaning performance}
\label{tab:cleaning_performance}
\begin{center}
\begin{small}
\begin{sc}
\begin{tabular}{lr}
\toprule
Metric & Value \\
\midrule
\multicolumn{2}{l}{\textit{Baseline}} \\
Original synthetic accuracy & 92.6\% \\
Original errors & 946 \\
\midrule
\multicolumn{2}{l}{\textit{Arena Correction}} \\
Arena accuracy & 95.0\% \\
Errors corrected & 786 / 946 (83.1\%) \\
Net improvement & +2.4\% \\
\midrule
\multicolumn{2}{l}{\textit{Bad Label Detection}} \\
Precision & 98.0\% \\
Recall & 95.2\% \\
F1 Score & 96.6\% \\
\bottomrule
\end{tabular}
\end{sc}
\end{small}
\end{center}
\vskip -0.1in
\end{table}

\paragraph{Per-Class Performance.}
Table~\ref{tab:per_class} presents per-label metrics. The arena performs best on CORRECT (F1 = 97.3\%), followed by UNKNOWN (F1 = 94.5\%). INCORRECT labels prove most challenging (F1 = 89.2\%), primarily due to lower recall---the arena tends to classify borderline INCORRECT cases as UNKNOWN, a conservative behavior preferable in production where false negatives are less costly than false positives.

\begin{table}[t]
\caption{Per-class performance metrics (averaged across languages)}
\label{tab:per_class}
\begin{center}
\begin{small}
\begin{sc}
\begin{tabular}{lccc}
\toprule
Label & Precision & Recall & F1-Score \\
\midrule
CORRECT & 96.2\% & 98.5\% & 97.3\% \\
INCORRECT & 91.5\% & 87.0\% & 89.2\% \\
UNKNOWN & 95.0\% & 94.1\% & 94.5\% \\
\bottomrule
\end{tabular}
\end{sc}
\end{small}
\end{center}
\vskip -0.1in
\end{table}

\paragraph{Performance by Original Label Type.}
Stratifying by original synthetic label reveals where the arena adds most value (Table~\ref{tab:by_original_type}). Originally CORRECT labels already achieve 97.3\% accuracy, with modest arena improvement (+1.2\%). Originally INCORRECT labels show the largest gap: synthetic accuracy is only 79.3\%, but the arena recovers to 88.7\% (+9.4\%), demonstrating particular effectiveness at correcting the most error-prone category.

\begin{table}[t]
\caption{Arena accuracy (\%) by original synthetic label type}
\label{tab:by_original_type}
\begin{center}
\begin{small}
\begin{sc}
\setlength{\tabcolsep}{4pt}
\begin{tabular}{lrrrr}
\toprule
Original & Count & Orig Acc & Arena Acc & $\Delta$ \\
\midrule
CORRECT & 5,980 & 97.3 & 98.5 & +1.2 \\
INCORRECT & 2,240 & 79.3 & 88.7 & +9.4 \\
UNKNOWN & 4,506 & 92.8 & 93.6 & +0.8 \\
\bottomrule
\end{tabular}
\end{sc}
\end{small}
\end{center}
\vskip -0.1in
\end{table}

\paragraph{Agreement Level and Accuracy.}
Accuracy correlates strongly with agreement level: unanimous consensus (36\% of products) yields 100\% accuracy, very high agreement ($>$85\%) yields 98.8\%, while low-agreement cases ($<$50\%, 1.4\% of products) achieve only 65.1\%, suggesting these warrant human review. Full stratification is provided in Appendix~\ref{appendix:results_detail}.

\paragraph{Label Quality.}
The 92.6\% baseline treats agreement cases as correct by construction; correcting for their triage-measured residual error (2.37\%) puts the synthetic labels at 90.5\% unconditionally, and the released labels at 97.9\% after adjudication (95\% CI [96.7\%, 99.0\%])---a silver rather than gold standard~\citep{northcutt2021pervasive}. See Appendix~\ref{appendix:quality}.

\begin{figure}[t]
    \centering
    \includegraphics[width=\columnwidth]{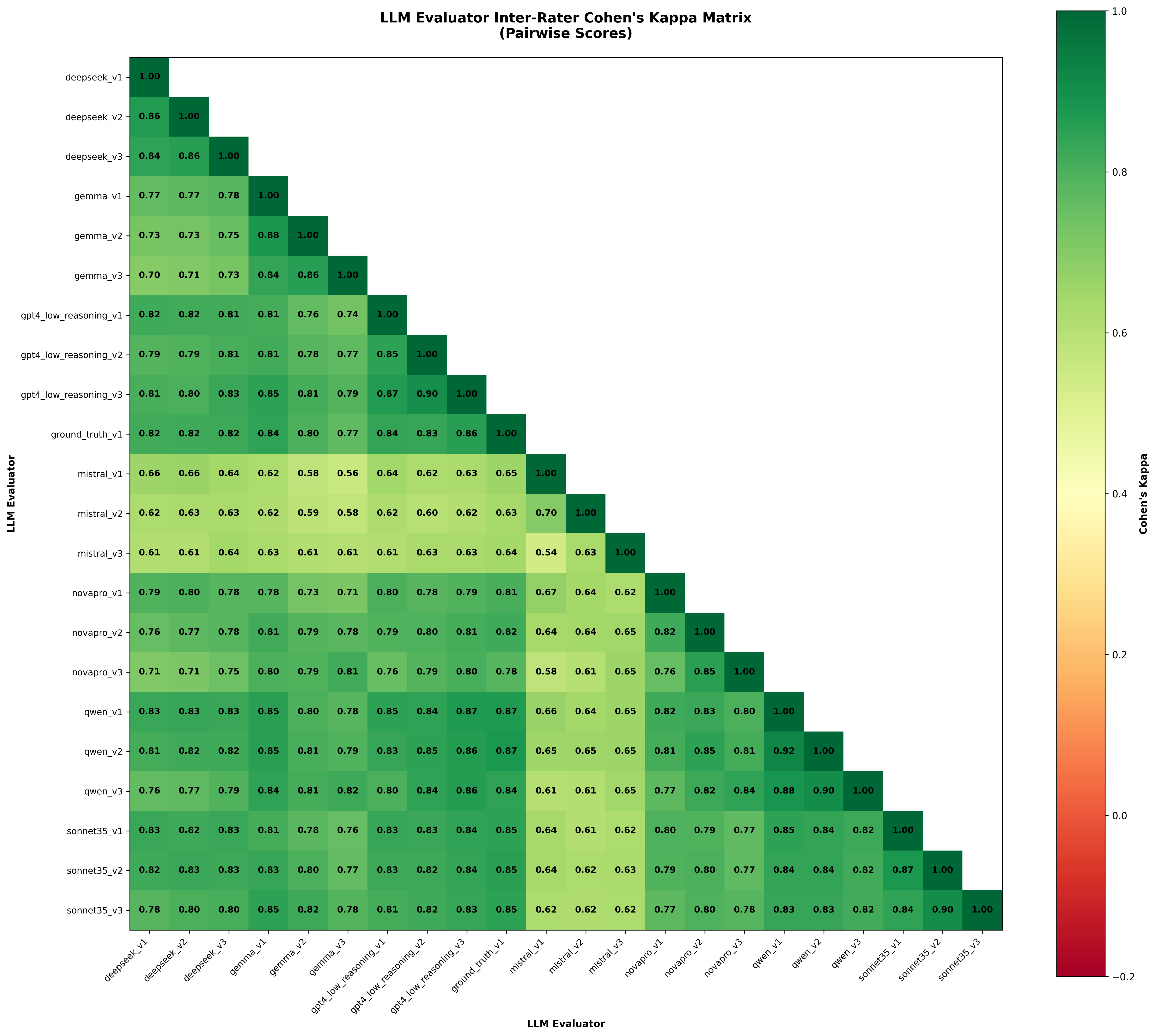}
    \caption{Pairwise Cohen's $\kappa$ between judge configurations (German). Within-family agreement is high (diagonal blocks), while cross-family diversity drives ensemble accuracy. Patterns hold across all languages (Appendix~\ref{appendix:heatmaps}).}
    \label{fig:german_cohen_heatmap}
\end{figure}

\paragraph{Additional Analyses.}
No individual judge configuration outperformed the ensemble, and accuracy varied more across model families than across prompts: at three judges, three families with one prompt each (91.8\%) beats one family with three prompts (89.9\%). The 21 configurations are thus not 21 independent signals---mean pairwise error correlation is 0.36, implying $\sim$2.6 effective votes~\citep{kohli2026nine}---matching the clustering in Figure~\ref{fig:german_cohen_heatmap}. The production-critical error is a false CORRECT, asserting an unsupported value: 235 of 630 errors (37.3\%), the rest being largely conservative abstentions (INCORRECT $\rightarrow$ UNKNOWN, 27.5\%). Italian shows the highest fix rate (90.2\%) and French the lowest (76.1\%). Full details are in Appendix~\ref{appendix:results_detail}.

\paragraph{Cost Analysis.}
Total validation cost was \$290.50 for 12,726 products (\$22.83 per 1,000 products, 267,246 API calls).\footnote{Pricing based on commercial API rates as of June 2025.} Detailed breakdowns are in Appendix~\ref{appendix:costs}.

\paragraph{The SynthAVE Dataset.}
The resulting dataset comprises 12,726 products spanning 229 product categories, 792 attributes, and 4 languages (47.7\% CORRECT, 15.8\% INCORRECT, 36.5\% UNKNOWN), with estimated label accuracy of 97.9\%. Labels are either expert-adjudicated (the 1,416 disagreement cases) or accepted on arena--pipeline consensus and validated by stratified audit (Appendix~\ref{appendix:triage}). Unlike MAVE~\citep{yang2021maveproductdatasetmultisource}, SynthAVE provides explicit UNKNOWN labels, multilingual coverage, and a documented validation protocol with a quantified residual error rate. The dataset and per-judge predictions will be publicly released.

%% file: sections/8_conclusion.tex
\section{Conclusion}
\label{sec:conclusion}

We presented SynthAVE, a large-scale benchmark and validation framework for synthetic e-commerce data. Building on prior work in synthetic data generation~\citep{negri2025attribute}, we introduced an LLM-arena framework that enables scalable quality assurance through multi-model evaluation.

Our evaluation yields three key findings. First, the LLM arena achieves 95\% agreement with expert evaluation (Cohen's $\kappa$ = 0.92) across 12,726 products, 229 product categories, 792 attributes, and 4 languages. Second, diversity rather than redundancy drives accuracy: judges agree with one another only substantially (Fleiss' $\kappa$ = 0.76, a diversity diagnostic rather than a shortfall), yet the ensemble outperforms any single model, with model families mattering more than prompts. Third, expert triage of 400 agreement cases shows the majority panel is overturned only 3\% of the time (0\% for unanimous cases), putting the estimated quality of the released labels at 97.9\%---a silver standard. The arena corrects 83.1\% of synthetic labeling errors at \$0.02 per product, reserving human review for panel--pipeline disagreements.

These results support adopting multi-LLM evaluation protocols that incorporate prompt and model diversity, moving beyond single-model approaches. The SynthAVE dataset will be publicly released to support future research in scalable data quality assurance for attribute-value verification.

%% file: sections/7_limitations.tex
\section*{Limitations}
\label{sec:limitations}

Several limitations warrant discussion. First, our evaluation covers four European languages with shared Latin script; generalization to languages with different scripts (e.g., Chinese, Arabic), morphological structures, or lower LLM proficiency requires additional validation. The framework was also evaluated only on e-commerce attribute verification---a task with relatively objective ground truth---and not on an external benchmark. MAVE~\citep{yang2021maveproductdatasetmultisource} is the closest candidate but is English-only, targets extraction rather than verification, and lacks explicit UNKNOWN labels, so transfer beyond SynthAVE remains open.

Second, the framework depends on the availability of diverse, high-capability LLMs, and specific judge configurations may require updating as model families evolve, retire, or change pricing. That some families substantially underperform others (accuracy ranging from 76\% to 93\%) raises the question of minimum capability thresholds for judge inclusion. We also aggregate by simple majority vote, a deliberately transparent baseline; given the clustered reliability structure documented in Appendix~\ref{appendix:panel}, reliability-weighted aggregation that down-weights redundant within-family judges is a natural extension we did not explore. Since the panel supplies roughly 2.6 effective votes rather than 21, gains from simply adding judges from existing families should be expected to be small.

Third, our ground truth construction relies on a disagreement-based annotation strategy in which human review was triggered only when the LLM ensemble disagreed with the synthetic label. Consequently 88.9\% of the benchmark was accepted without direct adjudication, and for those cases the reference labels are not independent of the arena's own judgments. Expert triage of 400 stratified agreement cases bounds the residual error at 3.0\% (0\% for unanimous cases), giving an estimated 97.9\% accuracy for the released labels (Appendix~\ref{appendix:quality})-- \textit{silver} rather than \textit{gold}-standard reliability, so comparisons between systems whose scores differ by less than roughly two points should be treated as unresolved~\citep{northcutt2021pervasive}. Adjudication of the disagreement cases rested on a single expert. A second annotator from a different team working in a similar catalog area independently re-labelled all 400 audit rows (Appendix~\ref{appendix:triage}), reproducing the first pass at Cohen's $\kappa = 0.98$ and a slightly more conservative overturn rate against the arena majority. Two caveats attach to this figure: the annotators share the general catalog domain and are expected to agree more than fully external experts, and re-annotating the full 1{,}416 disagreement cases would be required to bound annotator error on that stratum specifically.

Finally, while the framework is designed for offline validation rather than real-time inference, the latency of running 21 judge configurations may still be prohibitive for rapid iteration cycles. Future work could explore distilling ensemble judgments into smaller, specialized models for faster evaluation.

%% file: sections/appendix_triage.tex
\section{Human Triage of Agreement Cases}
\label{appendix:triage}

To validate the assumption that agreement between the synthetic generation pipeline and the LLM arena majority vote implies label correctness, an expert annotator independently triaged a stratified random sample of 400 agreement cases. The sample was balanced across locales (100 per language) and stratified by agreement level (100 unanimous, 300 mixed-agreement).

For each sample, the annotator reviewed the product text, attribute-value pair, and majority vote label, then recorded whether they \textit{agree} with the majority, \textit{disagree} (i.e., would overturn it), or are \textit{unsure}. The audit returned 382 agreements, 12 disagreements, and 6 unsure judgements.

\paragraph{Handling of Unsure Cases.}
Only clear disagreements are counted as overturns. Because the treatment of the six unsure cases is a modelling choice rather than an observation, we report all three defensible conventions: \textbf{3.0\%} (12/400), counting them as non-overturns, which is the figure used in the main text; \textbf{3.0\%} (12/394), excluding them from numerator and denominator alike; and \textbf{4.5\%} (18/400), the conservative upper bound obtained by counting every unsure case as an overturn. The conventions agree to within 1.5 percentage points, so no conclusion in this paper depends on the choice. Substituting the upper bound would raise the estimated residual error of the released labels (Appendix~\ref{appendix:quality}) from 2.1\% to 3.0\%, leaving the silver-standard characterisation intact.

\paragraph{Overall Results.}
The annotator disagreed with the majority vote in 12 out of 400 cases (3.0\%), with a 95\% confidence interval of [1.3\%, 4.7\%] (margin of error $\pm$1.7\%).

\begin{table}[h]
\caption{Expert triage overturn rates by slice. CI values are percentages.}
\label{tab:triage_results}
\begin{center}
\begin{small}
\begin{sc}
\setlength{\tabcolsep}{3pt}
\begin{tabular}{lrrr}
\toprule
Slice & Rate & 95\% CI & $n$ \\
\midrule
Overall & 3.0\% & [1.3, 4.7] & 400 \\
\midrule
\multicolumn{4}{l}{\textit{By sample kind}} \\
\quad Mixed     & 4.0\% & [1.8, 6.2] & 300 \\
\quad Unanimous & 0.0\% & ---        & 100 \\
\midrule
\multicolumn{4}{l}{\textit{By majority label}} \\
\quad UNKNOWN   & 5.0\% & [1.6, 8.3] & 161 \\
\quad CORRECT   & 2.2\% & [0.1, 4.4] & 181 \\
\quad INCORRECT & 0.0\% & ---        &  58 \\
\bottomrule
\end{tabular}
\end{sc}
\end{small}
\end{center}
\end{table}

\paragraph{Breakdown by Locale.}
Table ~\ref{tab:triage_locale} shows the expert triage overturn rates by locale.

\begin{table}[h]
\caption{Expert triage overturn rates by locale}
\label{tab:triage_locale}
\begin{center}
\begin{small}
\begin{sc}
\begin{tabular}{lrrr}
\toprule
Locale & Overturns & Rate & $n$ \\
\midrule
IT & 5 & 5.0\% & 100 \\
DE & 3 & 3.0\% & 100 \\
ES & 3 & 3.0\% & 100 \\
FR & 1 & 1.0\% & 100 \\
\bottomrule
\end{tabular}
\end{sc}
\end{small}
\end{center}
\end{table}

\paragraph{Key Findings.}
\begin{itemize}
    \item The majority panel achieves a 97\% accuracy rate (12/400 overturns) with a tight confidence interval of [1.3\%, 4.7\%].
    \item Unanimous cases have a 0\% overturn rate (0/100): when all 21 judges agree, the annotator found nothing to correct in our sample.
    \item UNKNOWN is the least reliable majority label at 5.0\% overturn rate, compared to CORRECT at 2.2\% and INCORRECT at 0.0\%. This is consistent with UNKNOWN being the conservative vote that can mask determinable cases.
    \item All 12 overturns came from mixed-agreement samples (4.0\% rate); no unanimous sample was overturned.
    \item Locale differences range from 1.0\% (FR) to 5.0\% (IT), but with $n$=100 per locale the margins of error (2--4\%) overlap substantially. Distinguishing locale-level effects would require $\sim$1,000+ samples per locale.
\end{itemize}

\paragraph{On the 0\% Unanimous Overturn Rate.}
One caveat deserves emphasis. For agreement cases the reference label \textit{is} the original synthetic label by construction, so a unanimous panel that matches it cannot disagree with the reference by definition. What makes the 0\% figure informative is the audit itself: an independent annotator, working from the product text rather than from the panel's output, found nothing to correct in 100 unanimous samples. The number should therefore be read as evidence that unanimity is a strong confidence signal, not as a measured classification accuracy of 100\%.

\paragraph{Second Expert Evaluation.}
To calibrate the reliability of the reference labels themselves, a second annotator independently re-labelled all 400 audit samples. Both annotators are researchers on this work with multiple years of experience on multilingual product-catalog attribute problems, drawn from two different teams operating in adjacent areas of catalog attribute research. The second annotator worked from a blinded interface in which the first annotator's label, the arena majority vote and the original triage decision were hidden. The annotators do not share day-to-day working conventions, but they do share the general catalog domain and the attribute taxonomy the benchmark is built on, so this study should be read as agreement between two experienced practitioners of the domain rather than as agreement with a fully external annotator.

\begin{table}[h]
\caption{Second-expert audit: agreement with the first annotator}
\label{tab:second_expert}
\begin{center}
\begin{small}
\begin{sc}
\setlength{\tabcolsep}{3.5pt}
\begin{tabular}{lrrr}
\toprule
Slice & Raw Agreement & Cohen $\kappa$ & Disagr. \\
\midrule
DE  & 98.0\% & 0.968 & 2 \\
ES  & 98.0\% & 0.968 & 2 \\
FR  & 99.0\% & 0.984 & 1 \\
IT  & 100.0\% & 1.000 & 0 \\
\midrule
\textbf{All (400)} & \textbf{98.8\%} & \textbf{0.980} & \textbf{5} \\
\bottomrule
\end{tabular}
\end{sc}
\end{small}
\end{center}
\end{table}

Across the 400 audit rows the two annotators agree on 395, giving Cohen's $\kappa = 0.98$ (95\% CI $[0.96,\,1.00]$), which sits in the almost-perfect band of \citet{landis1977measurement}. Applying the second annotator's judgements against the arena majority yields an overturn rate of 5/400 = 1.2\% overall, with 0/100 on unanimous samples and 5/300 = 1.7\% on mixed-agreement samples---consistent with, and slightly more conservative than, the 4.0\% mixed-sample overturn rate reported for the first pass. The first-pass 3.0\% figure is therefore reproducible, and both annotators would classify the audit set at 98--99\% accuracy.

The five human--human disagreements all occurred in mixed-agreement samples; four out of the five involve the arena's INCORRECT or UNKNOWN calls, where the first annotator sided with the arena and the second annotator would overturn it. Three cases (\texttt{washer\_type} and \texttt{thread\_type} in ES, and \texttt{guitar\_bridge\_system} in DE) flip an arena INCORRECT/UNKNOWN call to CORRECT under a broader reading of the value; one DE \texttt{max\_printspeed\_color} case flips INCORRECT to UNKNOWN; and one FR \texttt{section\_width} case flips a CORRECT call to UNKNOWN because the value is not explicitly stated in the text. Rather than silently override either annotator, we retain the first pass's labels as the released reference and treat the five residual disagreements as documented uncertainty; the point estimate of $\sim$97\% agreement-case accuracy therefore sits between the two annotators' independent overturn rates (1.2\% and 3.0\%), and the confidence interval in Appendix~\ref{appendix:quality} accommodates the difference.

\paragraph{Implications.}
The 3\% $\pm$ 1.7\% overturn rate supports the claim that the majority panel correctly labels $\sim$97\% of agreement cases. Combined with the 0\% overturn rate for unanimous cases, this validates the disagreement-based annotation strategy: agreement between the generation pipeline and a diverse 21-judge panel is a reliable signal of label correctness, with residual error concentrated in inherently ambiguous mixed-agreement cases. Appendix~\ref{appendix:quality} converts these rates into an unconditional estimate of the quality of the released labels.

%% file: sections/appendix_quality.tex
\section{Estimating Label Quality}
\label{appendix:quality}

The 92.6\% figure reported for the synthetic pipeline in Section~\ref{sec:results} is \textit{conditional}: it counts only errors surfaced in disagreement cases and treats agreement cases as correct by construction. Because the expert triage of Appendix~\ref{appendix:triage} measures a non-zero error rate inside the agreement stratum, we can replace this with an unconditional, stratified estimate.

\subsection{Stratified Estimator}

Partition the benchmark into the agreement stratum $A$ and the disagreement stratum $D$, with $P(A) = 88.9\%$ and $P(D) = 11.1\%$. The overall error rate is
$$\varepsilon = P(A)\,\varepsilon_A + P(D)\,\varepsilon_D.$$

Within $A$, we split further by consensus level. Unanimous samples make up 40.8\% of the stratum and show an observed overturn rate of 0\% (0/100 audited); mixed-agreement samples make up the remaining 59.2\% and show 4.0\% (12/300 audited). Hence
$$\varepsilon_A = 0.408 \times 0.0 + 0.592 \times 0.040 = 2.37\%.$$

\subsection{Before Expert Review}

Before adjudication, $\varepsilon_D$ is the rate at which the synthetic label is wrong among cases the arena contests, estimated at 66.8\% from the fixes applied to this stratum. This gives
$$\varepsilon = 0.889 \times 2.37\% + 0.111 \times 66.8\% = 9.5\%,$$
i.e. an unconditional synthetic-label accuracy of \textbf{90.5\%}, roughly two points below the conditional 92.6\% figure. The gap is entirely attributable to residual error in cases that the arena and the generator agreed upon and that were therefore never inspected.

\subsection{After Expert Review}

After expert adjudication, the disagreement stratum is human-labelled and contributes no residual error under the assumption that the adjudicating expert is correct, so $\varepsilon_D \approx 0$ and
$$\varepsilon = 0.889 \times 2.37\% + 0.111 \times 0\% = 2.1\%,$$
giving an estimated accuracy of \textbf{97.9\%} for the released labels. Propagating the binomial interval on the 12/300 mixed-sample overturn rate through the same weighting yields a 95\% confidence interval of [96.7\%, 99.0\%].

Two caveats apply. First, $\varepsilon_D \approx 0$ assumes the adjudicating expert is authoritative; with a single annotator we cannot bound this from human--human agreement, which is the principal limitation of the protocol. Second, the estimate inherits the sampling error of a 400-case audit, so the interval is wider than the point estimate suggests.

\begin{table}[h]
\caption{Estimated label quality before and after expert review}
\label{tab:quality_estimate}
\begin{center}
\begin{small}
\begin{sc}
\setlength{\tabcolsep}{3.5pt}
\begin{tabular}{lrrr}
\toprule
Stratum & Share & Err. Before & Err. After \\
\midrule
Agreement & 88.9\% & 2.37\% & 2.37\% \\
\quad unanimous & 40.8\% & 0.0\% & 0.0\% \\
\quad mixed & 59.2\% & 4.0\% & 4.0\% \\
Disagreement & 11.1\% & 66.8\% & $\approx$0\% \\
\midrule
\textbf{Overall error} & & \textbf{9.5\%} & \textbf{2.1\%} \\
\textbf{Accuracy} & & \textbf{90.5\%} & \textbf{97.9\%} \\
\bottomrule
\end{tabular}
\end{sc}
\end{small}
\end{center}
\end{table}

\subsection{Gold, Silver and Bronze Standards}

It is useful to place this figure on the conventional scale of reference-label quality. \textit{Gold standard} labels are manually verified and adjudicated by trusted domain experts, approaching 100\% accuracy at high cost. \textit{Silver standard} labels are produced with high confidence but without exhaustive human verification, typically by consensus among multiple annotators or a highly reliable automated procedure; they retain occasional noise. \textit{Bronze standard} labels come from heuristics, dictionary lookups, or weak models, and carry substantial noise in exchange for speed.

At an estimated 2.1\% residual error, SynthAVE sits at the upper end of the silver band: cleaner than the average of the widely used benchmarks audited by \citet{northcutt2021pervasive}, but short of the near-perfect agreement expected of a gold standard. This distinction matters for downstream use. \citet{northcutt2021pervasive} show that even small test-set label error rates can reorder model rankings, so benchmark consumers should treat differences between systems that fall within a couple of points as unresolved. Closing the remaining gap would require adjudicating the mixed-agreement portion of the agreement stratum, which our triage identifies as the sole location of residual error; the unanimous portion required no corrections in 100 audited samples.

%% file: sections/appendix_results_detail.tex
\section{Detailed Results}
\label{appendix:results_detail}

\subsection{Agreement Level and Accuracy}

Table~\ref{tab:agreement_accuracy} shows arena accuracy stratified by the proportion of judges agreeing on the majority vote.

\begin{table}[h]
\caption{LLM Arena accuracy by model agreement level}
\label{tab:agreement_accuracy}
\begin{center}
\begin{small}
\begin{sc}
\begin{tabular}{lrr}
\toprule
Agreement Level & Products & Accuracy \\
\midrule
Low ($<$50\%) & 172 & 65.1\% \\
Medium (50--70\%) & 1,198 & 79.5\% \\
High (70--85\%) & 1,398 & 91.1\% \\
Very High ($>$85\%) & 9,958 & 98.8\% \\
\midrule
Unanimous (100\%) & 4,612 & \textbf{100.0\%} \\
\bottomrule
\end{tabular}
\end{sc}
\end{small}
\end{center}
\end{table}

\subsection{Data Cleaning Performance by Language}

Table~\ref{tab:language_cleaning} presents detailed cleaning performance by language, including fix rates and regression rates.

\begin{table}[h]
\caption{Data cleaning performance by language}
\label{tab:language_cleaning}
\begin{center}
\begin{small}
\begin{sc}
\setlength{\tabcolsep}{3.5pt}
\begin{tabular}{lrrrr}
\toprule
Lang. & Orig Acc & Arena Acc & Fix & Regr. \\
\midrule
IT & 91.9\% & 96.4\% & 90.2\% & 3.0\% \\
DE & 93.8\% & 95.3\% & 93.2\% & 4.6\% \\
ES & 92.5\% & 94.7\% & 75.8\% & 3.8\% \\
FR & 92.1\% & 94.0\% & 76.1\% & 4.5\% \\
\bottomrule
\end{tabular}
\end{sc}
\end{small}
\end{center}
\end{table}

\subsection{Error Pattern Distribution}

Table~\ref{tab:error_patterns} breaks down the arena's 630 errors by confusion pattern. The dominant pattern is INCORRECT $\rightarrow$ UNKNOWN (27.5\%), indicating conservative abstention. Confusion involving UNKNOWN accounts for 80.5\% of all errors, while direct CORRECT $\leftrightarrow$ INCORRECT misclassification is rare (19.5\%).

\begin{table}[h]
\caption{Error pattern distribution (Human $\rightarrow$ Arena)}
\label{tab:error_patterns}
\begin{center}
\begin{small}
\begin{sc}
\begin{tabular}{lrr}
\toprule
Error Pattern & Count & \% of Errors \\
\midrule
INCORRECT $\rightarrow$ UNKNOWN & 173 & 27.5\% \\
UNKNOWN $\rightarrow$ CORRECT & 147 & 23.3\% \\
UNKNOWN $\rightarrow$ INCORRECT & 128 & 20.3\% \\
INCORRECT $\rightarrow$ CORRECT & 88 & 14.0\% \\
CORRECT $\rightarrow$ UNKNOWN & 59 & 9.4\% \\
CORRECT $\rightarrow$ INCORRECT & 35 & 5.6\% \\
\midrule
\textbf{Total Errors} & \textbf{630} & \textbf{100\%} \\
\bottomrule
\end{tabular}
\end{sc}
\end{small}
\end{center}
\end{table}

\subsection{Panel Size and Composition}
\label{appendix:panel}

Table~\ref{tab:panel_composition} reports majority-vote accuracy as a function of panel composition, averaging over every subset of a given size so that the numbers reflect a typical rather than a hand-picked panel. This sweep is recomputed from the per-judge prediction files released with the dataset; it reproduces the full-arena accuracy of Table~\ref{tab:arena_performance} to within 0.2 percentage points, and all comparisons below are between configurations evaluated on the same basis.

\begin{table}[h]
\caption{Accuracy by panel composition. Family subsets are averaged over all $\binom{7}{k}$ combinations.}
\label{tab:panel_composition}
\begin{center}
\begin{small}
\begin{sc}
\setlength{\tabcolsep}{3.5pt}
\begin{tabular}{llrrr}
\toprule
Composition & Judges & Mean & Min & Max \\
\midrule
\multicolumn{5}{l}{\textit{Family diversity ($k$ families $\times$ 3 prompts)}} \\
1 family & 3 & 89.9 & 81.0 & 92.9 \\
2 families & 6 & 92.7 & 90.4 & 93.8 \\
3 families & 9 & 93.7 & 91.9 & 94.5 \\
4 families & 12 & 94.2 & 93.4 & 94.7 \\
5 families & 15 & 94.6 & 94.1 & 95.0 \\
6 families & 18 & 94.9 & 94.7 & 95.0 \\
7 families & 21 & 95.2 & --- & --- \\
\midrule
\multicolumn{5}{l}{\textit{Prompt diversity (1 family, $p$ prompts)}} \\
1 prompt & 1 & 88.0 & 76.3 & 92.1 \\
2 prompts & 2 & 88.8 & 76.8 & 92.5 \\
3 prompts & 3 & 89.9 & 81.0 & 92.9 \\
\midrule
\multicolumn{5}{l}{\textit{Family diversity at fixed prompt}} \\
3 families & 3 & 91.8 & --- & 93.4 \\
7 families & 7 & 93.2 & --- & --- \\
\bottomrule
\end{tabular}
\end{sc}
\end{small}
\end{center}
\end{table}

Three conclusions follow. First, no single configuration reaches ensemble accuracy: the best individual judge attains 92.1\%, roughly three points below the full panel. Second, family diversity dominates prompt diversity at equal cost. Holding the judge budget at three, three families with one prompt each reaches 91.8\% while one family with three prompts reaches 89.9\%; adding prompt variants within a family buys under two points in total, whereas adding families buys more than five. Third, returns diminish but do not vanish: a nine-judge, three-family panel reaches 93.7\% on average and 94.5\% at best, within roughly one point of the full arena at 9\% of the cost (Appendix~\ref{appendix:costs}). For deployments where a one-point accuracy loss is acceptable, this is the configuration we would recommend; pruning prompts is safer than pruning families.

\subsection{Judge Dependence and Effective Votes}

Because the panel aggregates correlated judges, the nominal count of 21 overstates the independent evidence available. Following \citet{kohli2026nine}, we estimate the effective number of votes from the mean pairwise correlation $\bar{\rho}$ of judge error indicators as $n_{\text{eff}} = n / (1 + (n-1)\bar{\rho})$.

We measure $\bar{\rho} = 0.36$ across all 210 judge pairs, decomposing into 0.53 for within-family pairs and 0.34 for cross-family pairs. This yields $n_{\text{eff}} \approx 2.6$ effective votes from 21 configurations. The interpretation is not that the panel is ineffective---it outperforms every individual judge---but that its reliability should be attributed to the diversity of a handful of genuinely distinct error profiles rather than to the nominal panel size. The within- versus cross-family split also explains the composition results above: within-family judges are largely redundant, so the marginal value of a new prompt is small relative to that of a new family.

\subsection{Error Distribution by Category, Attribute and Language}
\label{appendix:error_slices}

The dominant production risk is a false CORRECT, in which the panel asserts a value the product text does not support. Table~\ref{tab:error_slices} reports the categories and attributes where the arena is least reliable.

\begin{table}[h]
\caption{Highest and lowest error rates by product category and attribute (slices with $n \geq 60$)}
\label{tab:error_slices}
\begin{center}
\begin{small}
\begin{sc}
\setlength{\tabcolsep}{3.5pt}
\begin{tabular}{lrr}
\toprule
Slice & $n$ & Error \\
\midrule
\multicolumn{3}{l}{\textit{Categories, highest error}} \\
WALL\_ART & 62 & 14.5\% \\
AIMING\_SCOPE\_SIGHT & 66 & 10.6\% \\
CLOCK & 66 & 10.6\% \\
BINOCULAR & 69 & 10.1\% \\
TABLE & 89 & 9.0\% \\
TELEVISION & 98 & 8.2\% \\
\midrule
\multicolumn{3}{l}{\textit{Categories, lowest error}} \\
KICK\_SCOOTER & 105 & 1.0\% \\
SWIMWEAR & 98 & 1.0\% \\
MICROPHONE & 60 & 0.0\% \\
\midrule
\multicolumn{3}{l}{\textit{Attributes, highest error}} \\
handle.material & 106 & 7.5\% \\
container.type & 100 & 6.0\% \\
number\_of\_pockets & 71 & 5.6\% \\
power\_source\_type & 76 & 5.3\% \\
display.type & 150 & 4.7\% \\
\bottomrule
\end{tabular}
\end{sc}
\end{small}
\end{center}
\end{table}

Errors concentrate where the attribute is a fine-grained physical property that product copy tends to describe loosely. \texttt{handle.material} and \texttt{container.type} are the clearest instances: descriptions mention a material or container in passing without specifying the one being asked about, and judges over-commit to CORRECT on the basis of a nearby but different mention. Decorative and optical categories (WALL\_ART, AIMING\_SCOPE\_SIGHT, BINOCULAR) show the same pattern for framing, mounting and coating attributes.

Representative false-CORRECT cases illustrate the mechanism. For a wall clock with attribute \texttt{alarm\_clock} and generated value \texttt{FALSE}, the text neither confirms nor denies an alarm function; the correct label is UNKNOWN, but the panel returned CORRECT. For a safety-glasses product with \texttt{frame.type} = ``flexible fit'', the description praises comfort without characterising the frame; again UNKNOWN, again asserted CORRECT. For a utility cart with \texttt{caster.type} = ``universal front wheels'', the text describes wheels without specifying caster type, and the panel confirmed a value the ground truth marks INCORRECT.

False CORRECT rates also vary by language, from 1.4\% in German to 2.5\% in Italian, tracking the overall per-language ordering in Table~\ref{tab:language_cleaning}. Practitioners deploying this pipeline should apply tighter thresholds, or route to human review, on attribute families of this kind rather than applying a uniform confidence policy across the catalog.

%% file: sections/appendix_dataset_stats.tex
\section{Dataset Statistics}
\label{appendix:dataset_stats}

This appendix provides comprehensive statistics for the SynthAVE test set we release with this paper.

\subsection{Overall Scale}

\begin{itemize}
    \item \textbf{Total products}: 12,726 unique products
    \item \textbf{Product categories}: 229 distinct types
    \item \textbf{Unique attributes}: 792 across all languages
    \item \textbf{Product-attribute combinations}: 2,607 unique pairs
    \item \textbf{Average entries per combination}: 4.9
    \item \textbf{Languages}: 4 (Spanish, French, Italian, German)
    \item \textbf{Products per language}: 3,007--3,501
    \item \textbf{Judge configurations}: 21 (7 models $\times$ 3 prompts)
\end{itemize}

\subsection{Label Distribution}

\begin{table}[h]
\caption{Overall label distribution (majority vote)}
\label{tab:label_distribution}
\begin{center}
\begin{small}
\begin{sc}
\begin{tabular}{lrr}
\toprule
Label & Count & \% \\
\midrule
CORRECT & 6,214 & 48.8 \\
UNKNOWN & 4,602 & 36.2 \\
INCORRECT & 1,910 & 15.0 \\
\midrule
Total & 12,726 & 100.0 \\
\bottomrule
\end{tabular}
\end{sc}
\end{small}
\end{center}
\end{table}

Table~\ref{tab:label_distribution} presents the overall label distribution based on majority vote across all judges.

\subsection{Product Category Distribution}

Table~\ref{tab:category_diversity} presents the top 20 product categories by volume.

\begin{table}[h]
\caption{Top 20 product categories by count (of 229 total)}
\label{tab:category_diversity}
\begin{center}
\begin{small}
\begin{sc}
\begin{tabular}{lrr}
\toprule
Category & Count & \% \\
\midrule
NOTEBOOK\_COMPUTER & 209 & 1.6 \\
CELLULAR\_PHONE & 190 & 1.5 \\
HANDBAG & 180 & 1.4 \\
SHOES & 140 & 1.1 \\
PRINTER & 138 & 1.1 \\
WATCH & 133 & 1.0 \\
CAMERA\_DIGITAL & 127 & 1.0 \\
CHAIR & 123 & 1.0 \\
SECURITY\_CAMERA & 122 & 1.0 \\
WEARABLE\_COMPUTER & 121 & 1.0 \\
DRESS & 118 & 0.9 \\
LIGHT\_FIXTURE & 115 & 0.9 \\
CABINET & 111 & 0.9 \\
KICK\_SCOOTER & 110 & 0.9 \\
BACKPACK & 110 & 0.9 \\
VIDEO\_PROJECTOR & 108 & 0.8 \\
BRA & 107 & 0.8 \\
PERSONAL\_COMPUTER & 107 & 0.8 \\
HEADPHONES & 107 & 0.8 \\
MONITOR & 106 & 0.8 \\
\midrule
\multicolumn{3}{l}{\textit{...209 additional categories}} \\
\bottomrule
\end{tabular}
\end{sc}
\end{small}
\end{center}
\end{table}

\subsection{Attribute Distribution}

Table~\ref{tab:attribute_diversity} presents the top 20 attributes by frequency.

\begin{table}[h]
\caption{Top 20 attributes by count (of 792 total)}
\label{tab:attribute_diversity}
\begin{center}
\begin{small}
\begin{sc}
\begin{tabular}{lrr}
\toprule
Attribute & Count & \% \\
\midrule
closure.type & 296 & 2.3 \\
frame.material & 290 & 2.3 \\
brand & 237 & 1.9 \\
outer.material & 185 & 1.5 \\
material & 170 & 1.3 \\
display.type & 163 & 1.3 \\
color & 132 & 1.0 \\
waist.size & 121 & 1.0 \\
battery.average\_life & 115 & 0.9 \\
neck.neck\_style & 115 & 0.9 \\
handle.material & 111 & 0.9 \\
container.type & 108 & 0.8 \\
top.material & 107 & 0.8 \\
chest.size & 106 & 0.8 \\
sleeve.type & 100 & 0.8 \\
display.size & 88 & 0.7 \\
special\_feature & 87 & 0.7 \\
size & 86 & 0.7 \\
weight\_capacity.maximum & 84 & 0.7 \\
metal\_type & 81 & 0.6 \\
\midrule
\multicolumn{3}{l}{\textit{...772 additional attributes}} \\
\bottomrule
\end{tabular}
\end{sc}
\end{small}
\end{center}
\end{table}

\subsection{Top Product-Attribute Combinations}

Table~\ref{tab:top_combinations} presents the most frequent product-attribute combinations in the dataset.

\begin{table}[h]
\caption{Top 20 product-attribute combinations (of 2,607 total)}
\label{tab:top_combinations}
\begin{center}
\begin{small}
\begin{sc}
\setlength{\tabcolsep}{3pt}
\begin{tabular}{llr}
\toprule
Product Type & Attribute & N \\
\midrule
RADIO & display.type & 27 \\
THERMOMETER & outer.material & 25 \\
ELEC.\_CIGARETTE & battery.power & 25 \\
BLANKET & blanket\_form & 24 \\
SUITCASE & handle.type & 24 \\
ENVELOPE & closure.type & 23 \\
CLOTHES\_RACK & frame.material & 23 \\
AIMING\_SCOPE & has\_night\_vision & 23 \\
WALKING\_STICK & base.material & 22 \\
SEASONING & container.type & 22 \\
HANDBAG & material & 22 \\
POWER\_BANK & battery.cell\_comp & 22 \\
INCONT.\_PROT. & incont.\_prot.\_type & 21 \\
OUTBUILDING & frame.material & 21 \\
WATCH\_BAND & watch\_band\_design & 21 \\
PICTURE\_FRAME & frame.material & 21 \\
SELF\_BAL.\_VEH. & wheel.size & 20 \\
WALLET & wallet\_comp.\_type & 20 \\
WATCH & dial.color & 20 \\
SUNGLASSES & polarization\_type & 20 \\
\bottomrule
\end{tabular}
\end{sc}
\end{small}
\end{center}
\end{table}

\subsection{Per-Language Statistics}

Table~\ref{tab:language_detailed} provides detailed per-language statistics including unique product types, attributes, and combinations.

\begin{table}[h]
\caption{Detailed statistics by language}
\label{tab:language_detailed}
\begin{center}
\begin{small}
\begin{sc}
\setlength{\tabcolsep}{3.5pt}
\begin{tabular}{lrrrr}
\toprule
& ES & FR & IT & DE \\
\midrule
Total entries & 3,159 & 3,501 & 3,007 & 3,059 \\
Product types & 227 & 229 & 225 & 229 \\
Attributes & 636 & 675 & 609 & 673 \\
Combinations & 1,443 & 1,505 & 1,267 & 1,483 \\
\midrule
\multicolumn{5}{l}{\textit{Majority Vote Counts}} \\
CORRECT & 1,536 & 1,710 & 1,477 & 1,491 \\
INCORRECT & 482 & 553 & 437 & 438 \\
UNKNOWN & 1,141 & 1,238 & 1,093 & 1,130 \\
\bottomrule
\end{tabular}
\end{sc}
\end{small}
\end{center}
\end{table}

\subsection{Ground Truth Distribution}

Reference labels come from expert adjudication for the disagreement cases and from the accepted synthetic label for the agreement cases (Section~\ref{sec:framework}). Table~\ref{tab:ground_truth} shows the resulting distribution.

\begin{table}[h]
\caption{Ground truth label distribution (human-verified)}
\label{tab:ground_truth}
\begin{center}
\begin{small}
\begin{sc}
\begin{tabular}{lrr}
\toprule
Ground Truth & Count & \% \\
\midrule
CORRECT & 6,073 & 47.7 \\
UNKNOWN & 4,645 & 36.5 \\
INCORRECT & 2,008 & 15.8 \\
\midrule
Total & 12,726 & 100.0 \\
\bottomrule
\end{tabular}
\end{sc}
\end{small}
\end{center}
\end{table}

\subsection{Effect of Validation on the Label Mix}
\label{appendix:label_shift}

Table~\ref{tab:label_shift} compares the original synthetic labels with the final reference labels. Validation is a mild correction rather than a redistribution: CORRECT gains 93 samples (+0.7 points), INCORRECT loses 232 ($-$1.8), and UNKNOWN gains 139 (+1.1). This is expected, because the generation protocol fixes a target label for each sample in advance, so validation corrects execution errors rather than re-balancing the design. The largest movement is out of INCORRECT, consistent with the per-class results in Appendix~\ref{appendix:results_detail}, where INCORRECT is both the least accurate original label and the class the arena most often revises towards UNKNOWN.

\begin{table}[h]
\caption{Original synthetic vs.\ final reference label distribution}
\label{tab:label_shift}
\begin{center}
\begin{small}
\begin{sc}
\setlength{\tabcolsep}{3.5pt}
\begin{tabular}{lrrrr}
\toprule
Label & Original & \% & Final & \% \\
\midrule
CORRECT & 5,980 & 47.0 & 6,073 & 47.7 \\
INCORRECT & 2,240 & 17.6 & 2,008 & 15.8 \\
UNKNOWN & 4,506 & 35.4 & 4,645 & 36.5 \\
\midrule
Total & 12,726 & 100.0 & 12,726 & 100.0 \\
\bottomrule
\end{tabular}
\end{sc}
\end{small}
\end{center}
\end{table}

\textbf{Practical utility.} Because the label mix is set by design, SynthAVE is intended as a diagnostic test set: per-class scores should be re-weighted by the label prior of the target application before being read as expected error rates. What transfers directly is the relative ordering of systems and their behaviour on the hard slices identified in Appendix~\ref{appendix:results_detail}.

%% file: sections/appendix_heatmaps.tex
\section{Inter-Judge Agreement Heatmaps by Language}
\label{appendix:heatmaps}

This appendix presents pairwise agreement heatmaps for all four languages. 
Figure~\ref{fig:german_cohen_heatmap} in the main text shows German (Cohen's $\kappa$). Below we provide the complete set for all languages and both 
agreement metrics.

\subsection{Cohen's Kappa Heatmaps}
\begin{figure}[h]
    \centering
    \includegraphics[width=\columnwidth]{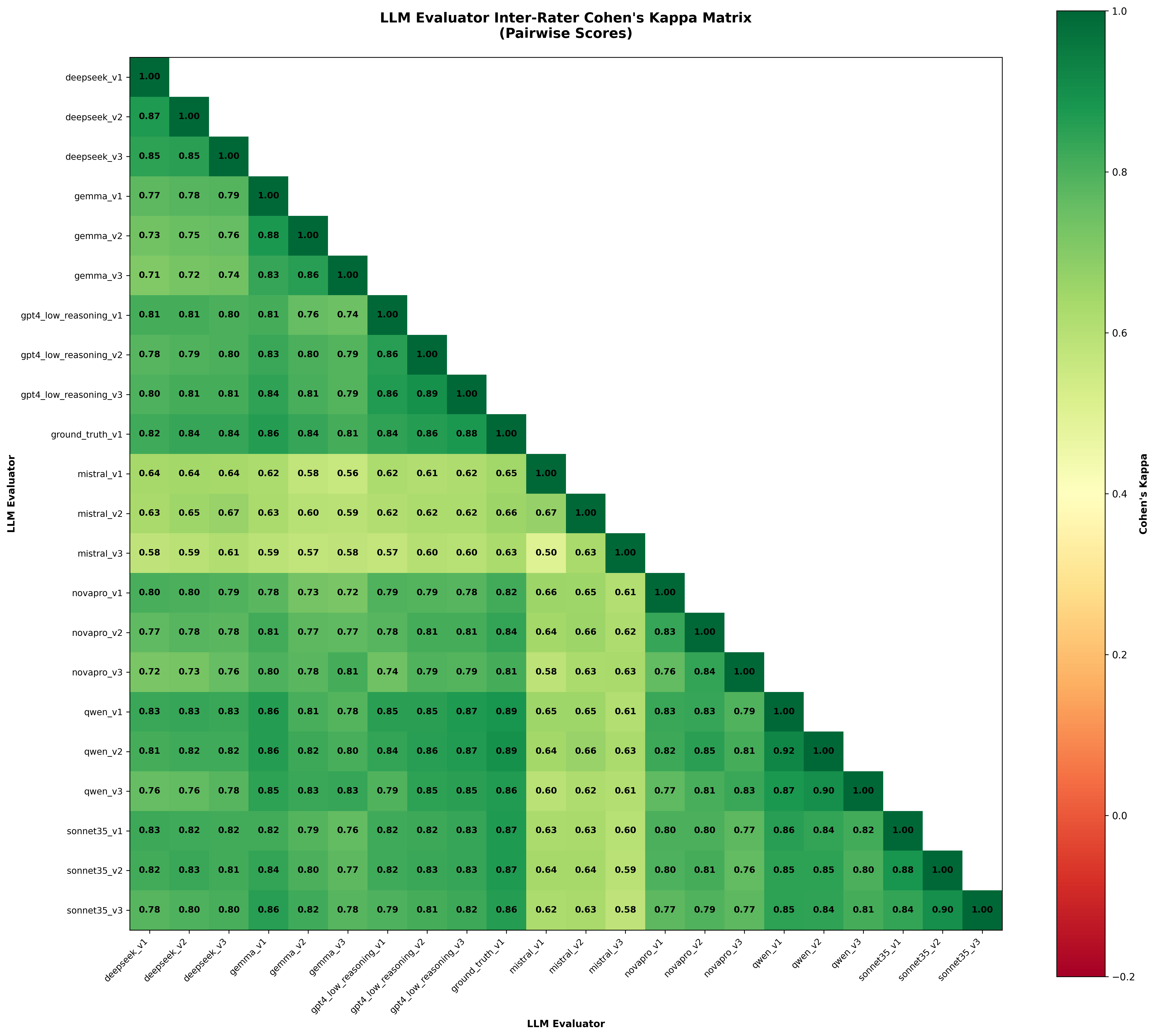}
    \caption{Pairwise Cohen's $\kappa$ between judge configurations for Italian. 
    Clustering patterns are consistent with other languages, with the same model 
    families forming high-agreement and low-agreement groups.}
    \label{fig:italian_cohen_heatmap}
\end{figure}

\begin{figure}[h]
    \centering
    \includegraphics[width=\columnwidth]{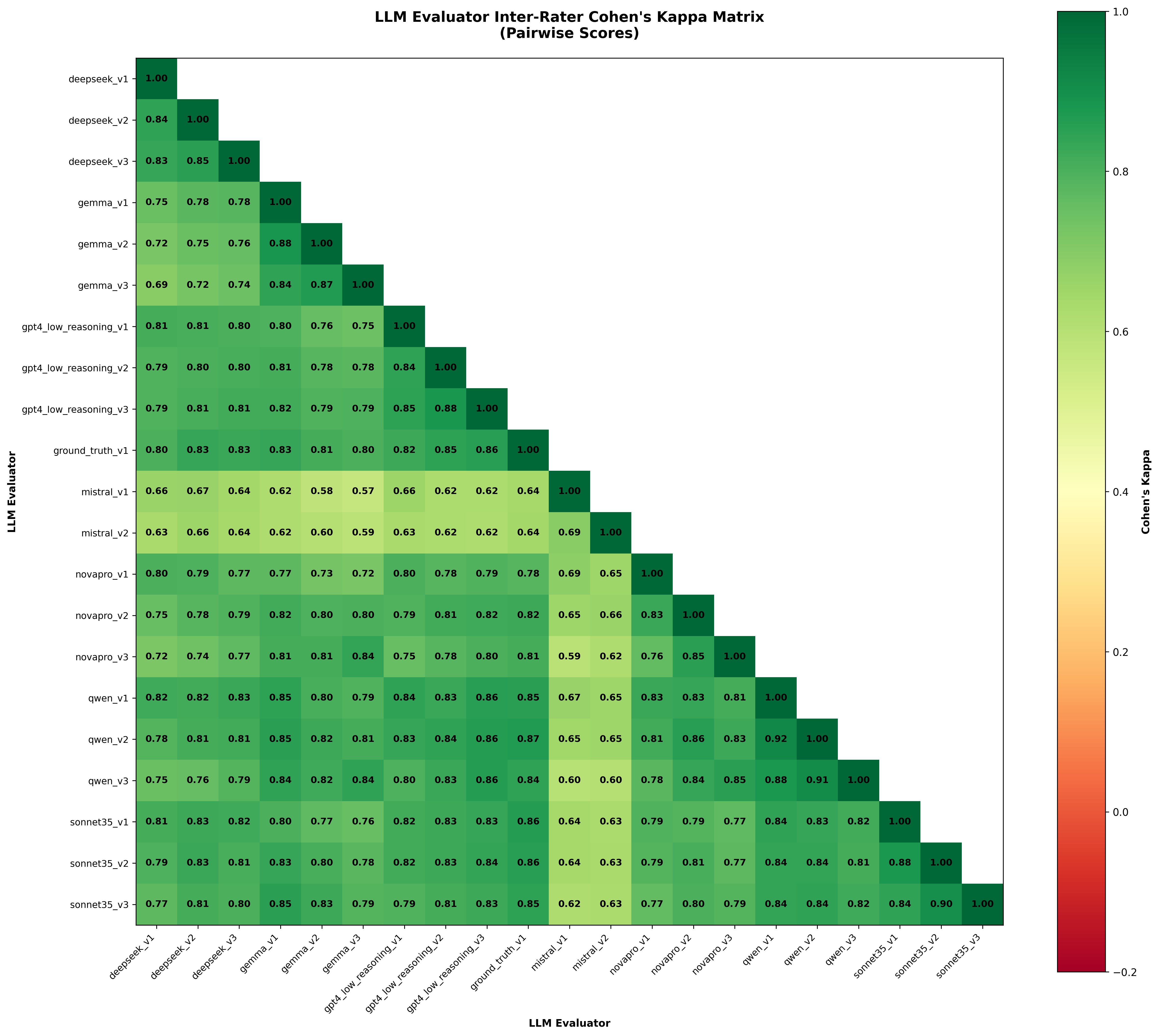}
    \caption{Pairwise Cohen's $\kappa$ between judge configurations for French. 
    Despite French showing the lowest overall arena accuracy (94.0\%), 
    inter-judge agreement patterns remain consistent with other languages.}
    \label{fig:french_cohen_heatmap}
\end{figure}

\begin{figure}[h]
    \centering
    \includegraphics[width=\columnwidth]{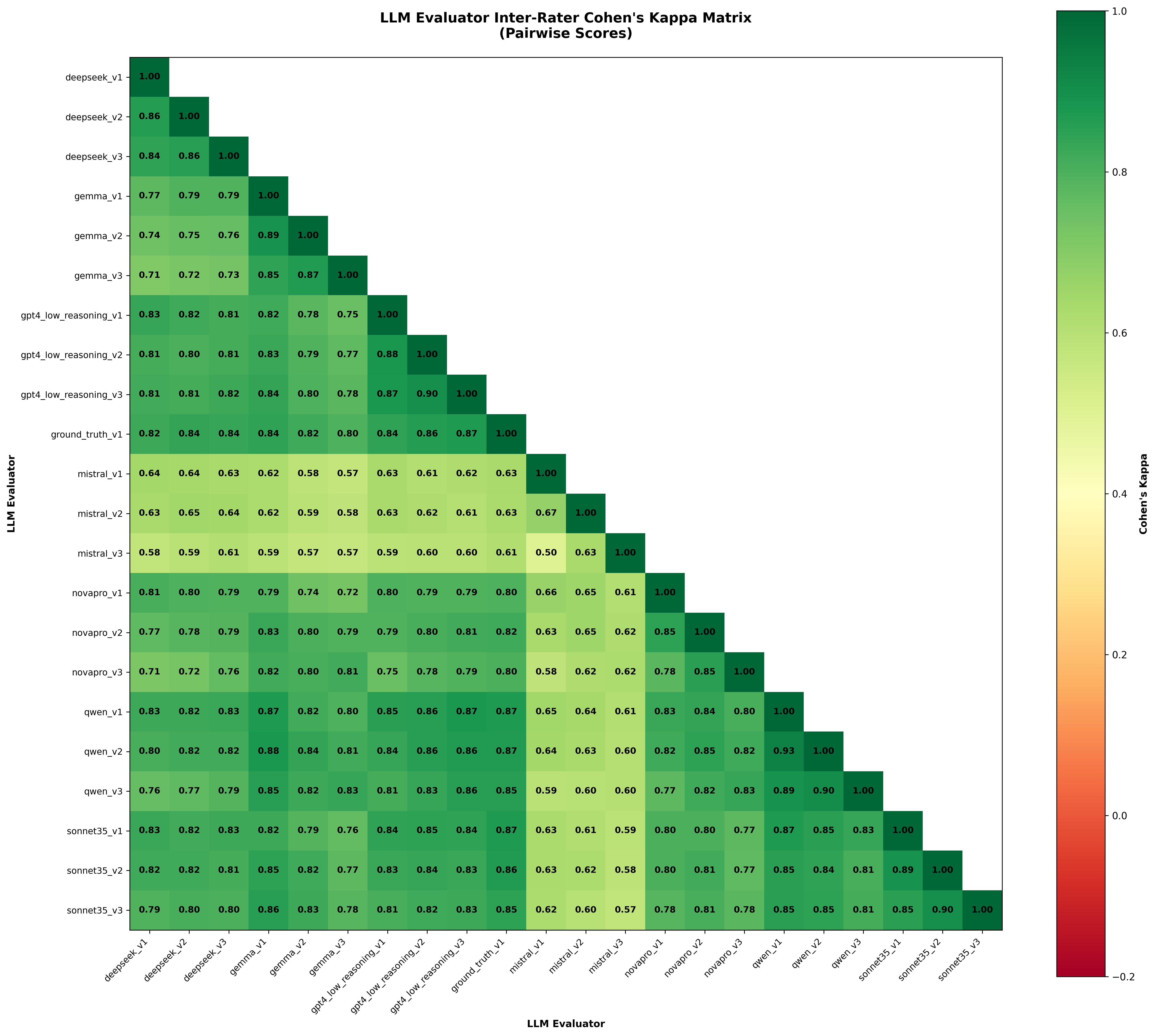}
    \caption{Pairwise Cohen's $\kappa$ between judge configurations for Spanish.}
    \label{fig:spanish_cohen_heatmap}
\end{figure}

Figures \ref{fig:italian_cohen_heatmap}, \ref{fig:french_cohen_heatmap}, \ref{fig:spanish_cohen_heatmap}

\subsection{Raw Agreement Heatmaps}
\begin{figure}[h]
    \centering
    \includegraphics[width=\columnwidth]{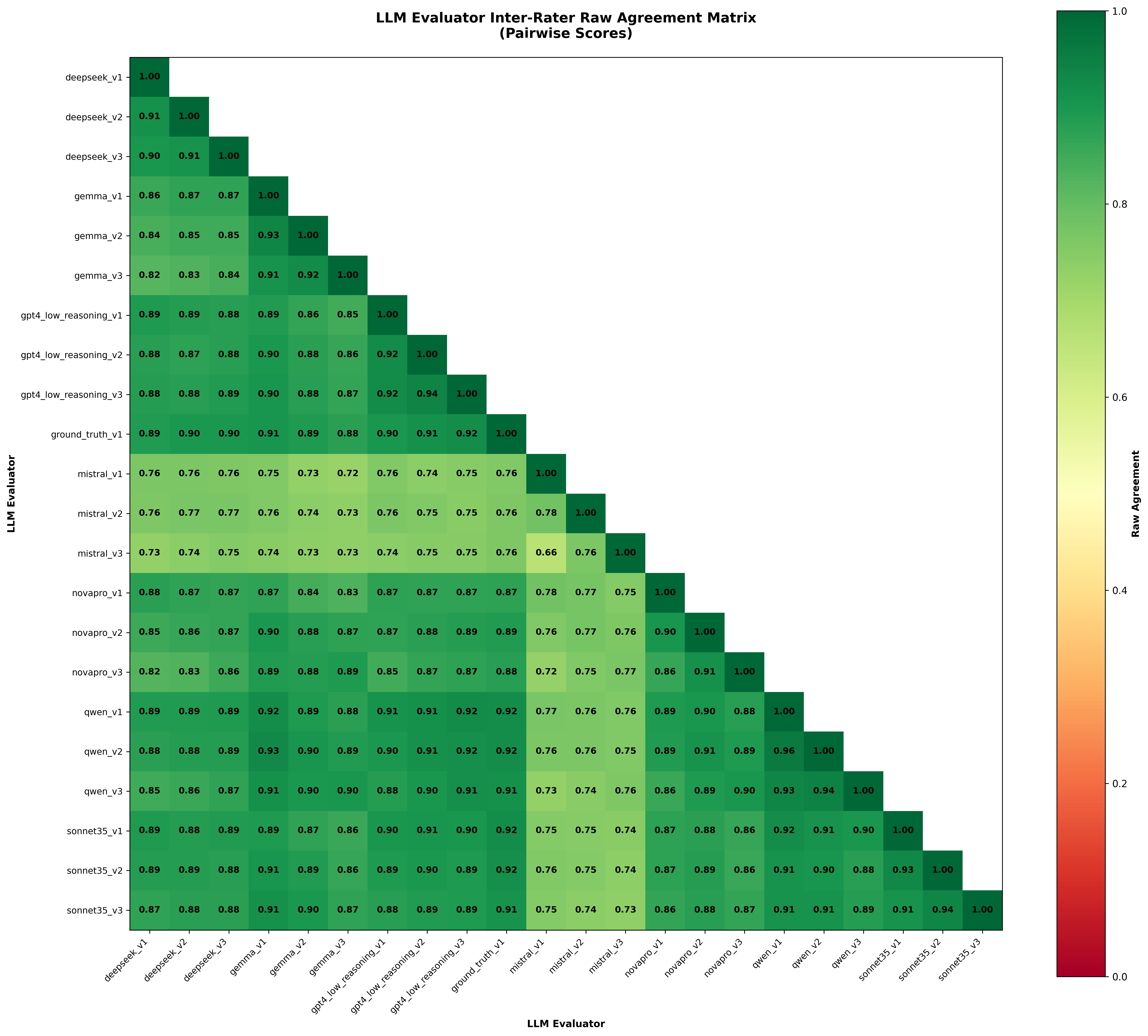}
    \caption{Pairwise raw agreement between judge configurations for Spanish. 
    Agreement ranges from 75\% to 96\% (within-family pairs).}
    \label{fig:spanish_agreement_heatmap}
\end{figure}

\begin{figure}[h]
    \centering
    \includegraphics[width=\columnwidth]{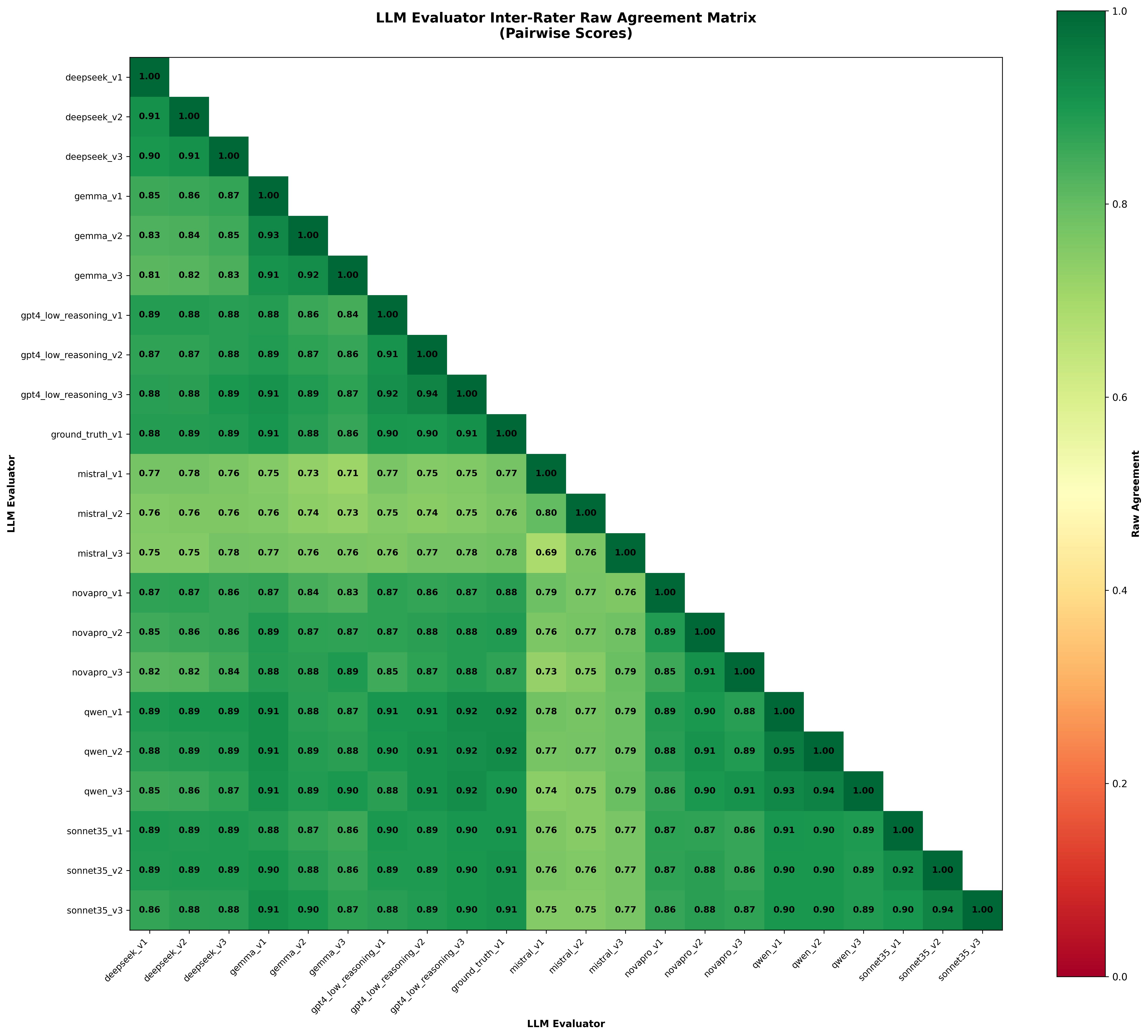}
    \caption{Pairwise raw agreement between judge configurations for German. 
    Agreement ranges from approximately 75\% (cross-family pairs with 
    lower-performing models) to 96\% (within-family pairs).}
    \label{fig:german_agreement_heatmap}
\end{figure}

\begin{figure}[h]
    \centering
    \includegraphics[width=\columnwidth]{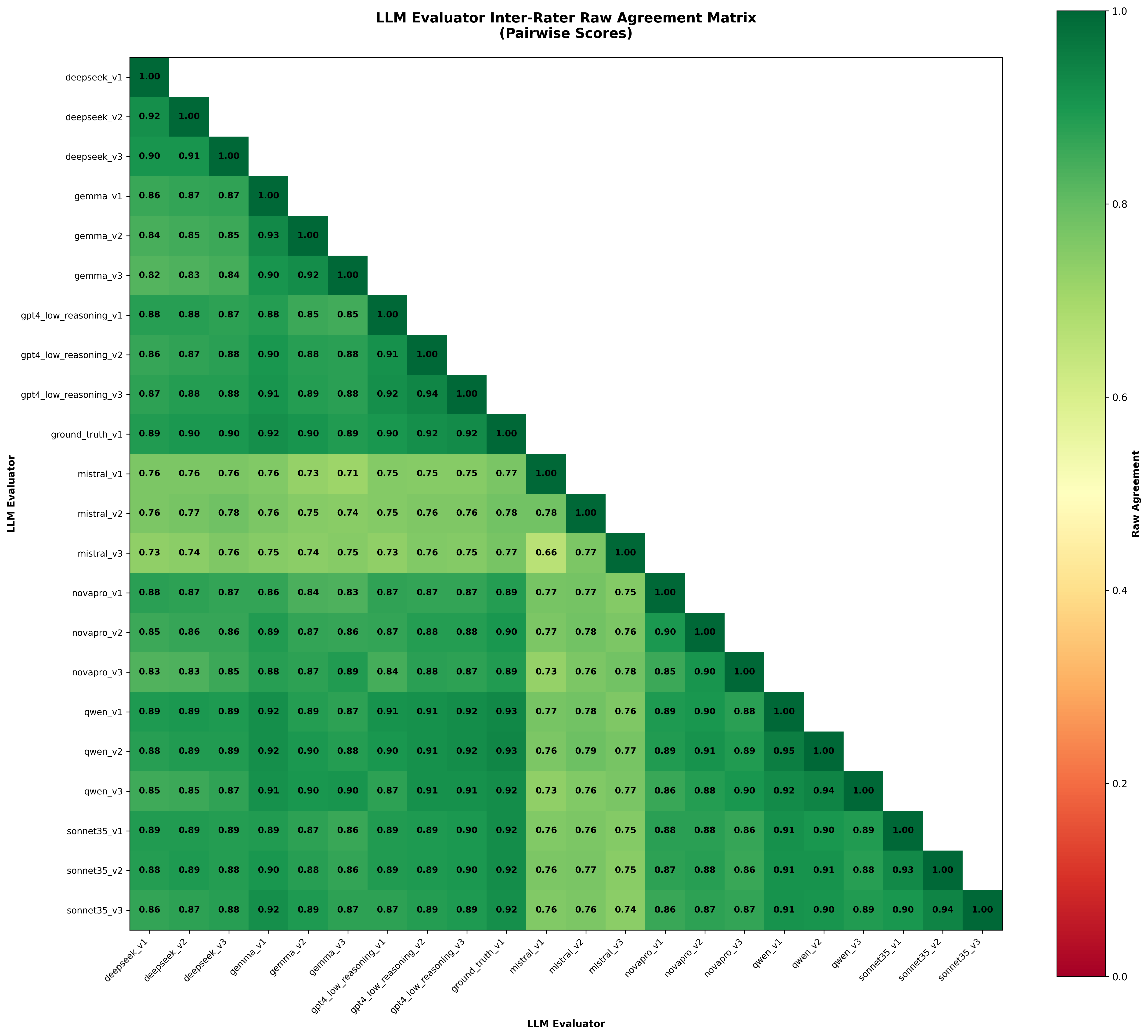}
    \caption{Pairwise raw agreement between judge configurations for Italian. 
    Italian shows the highest overall agreement, consistent with its 
    leading arena accuracy (96.4\%).}
    \label{fig:italian_agreement_heatmap}
\end{figure}

\begin{figure}[h]
    \centering
    \includegraphics[width=\columnwidth]{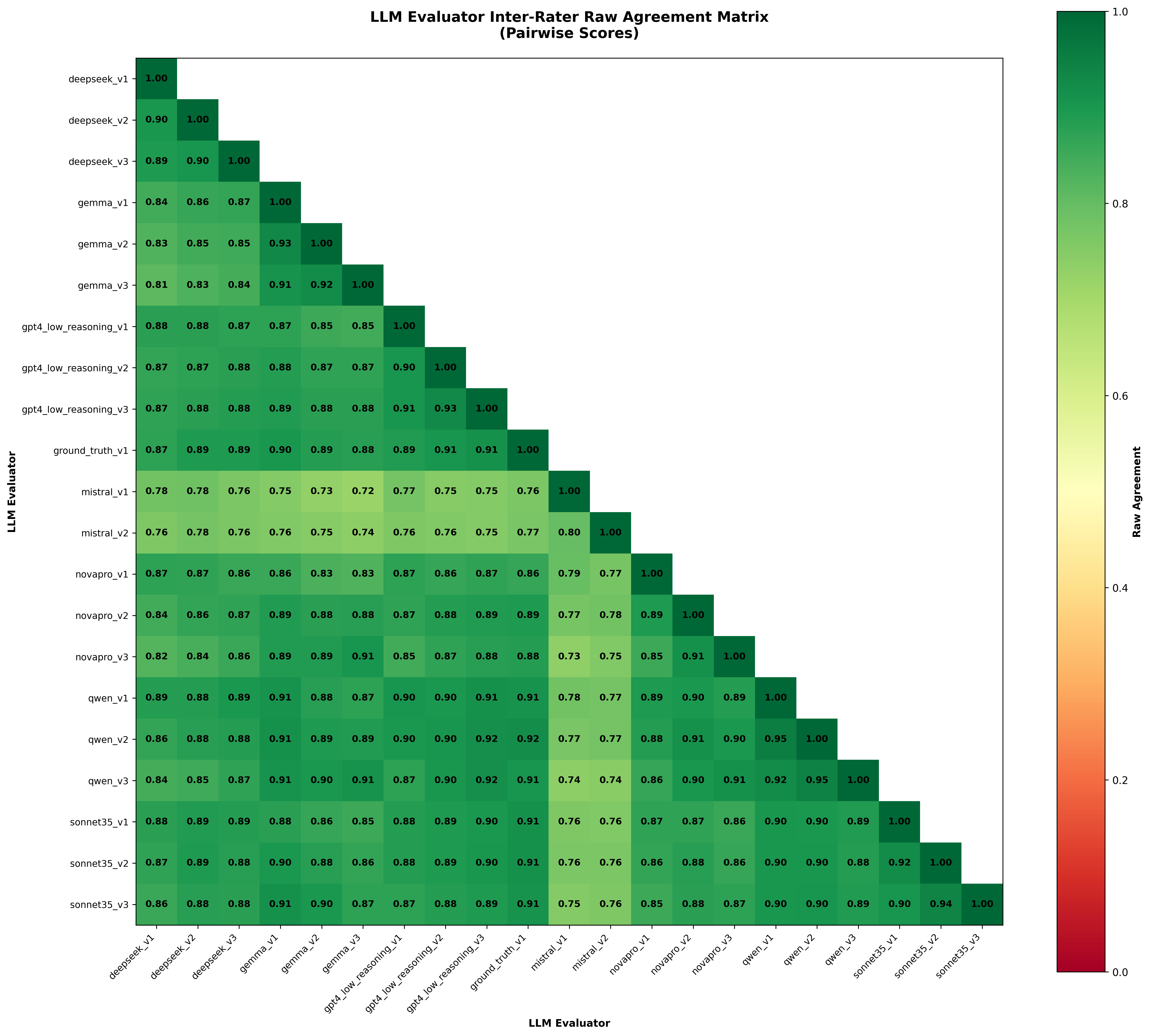}
    \caption{Pairwise raw agreement between judge configurations for French.}
    \label{fig:french_agreement_heatmap}
\end{figure}

Figures \ref{fig:spanish_agreement_heatmap}, \ref{fig:german_agreement_heatmap}, \ref{fig:italian_agreement_heatmap}, \ref{fig:french_agreement_heatmap}

%% file: sections/appendix_prompts.tex
\section{Evaluation Prompt Templates}
\label{appendix:prompts}

This appendix provides the complete prompt templates used for our LLM Arena evaluation across all three prompt versions. Each version maintains the same evaluation task (attribute consistency verification) while employing different structural approaches and reasoning strategies to ensure IID generation requirements.

\subsection{Prompt Version 1: Direct Instruction}

\begin{figure*}[t]
\begin{scriptsize}
\begin{verbatim}
You are an expert product attribute evaluator. Your task is to determine whether a generated attribute label is correct,
incorrect, or unknown based on the provided product information.

PRODUCT INFORMATION:
ID: {id}  |  Category: {category}  |  Title: {title}
Product Description: {description}
Bullet Points: {bullet_points}
ATTRIBUTE: {attribute}  |  GENERATED LABEL: {generated_label}

TASK: Evaluate whether the generated label "{generated_label}" is accurate for the attribute "{attribute}" based on the product
information provided. Note that some product information (description or bullet points) may not be available.

RESPONSE OPTIONS:
- CORRECT: The generated label accurately reflects the product's attribute based on the available information
- INCORRECT: The generated label is clearly wrong based on the product information
- UNKNOWN: Cannot determine if the label is correct or incorrect from the available product information
  (including cases where insufficient information is provided)

RESPONSE FORMAT: You must respond with exactly one of: "CORRECT", "INCORRECT", or "UNKNOWN".
After your choice, in a new line, provide a brief explanation (1-2 sentences) of your reasoning.

Example response format:
CORRECT
The product description explicitly mentions this attribute value.
\end{verbatim}
\end{scriptsize}
\caption{Prompt Version 1: Direct instruction format with minimal structure.}
\label{fig:prompt_v1}
\end{figure*}

\subsection{Prompt Version 2: XML-Structured with Examples}

\begin{figure*}[t]
\begin{scriptsize}
\begin{verbatim}
You are an e-Commerce data analyst specializing in product attribute verification and quality assurance.
Assess whether a generated attribute label correctly represents the specified product attribute based on available listing information.

<examples>
<example>
ID: B08N5WRWNW | Category: LUGGAGE
Title: <brandname> <makename> Hardside Expandable Luggage with Spinner Wheels, Brushed Anthracite, Carry-On 20-Inch
Description: Durable polycarbonate shell construction with scratch-resistant texture
Bullet Points: * 100% Polycarbonate shell * Four multidirectional spinner wheels * TSA-approved combination lock
Attribute Name: material_type | Generated Label: Polycarbonate
Your Response:
CORRECT
The description and bullet points explicitly confirm polycarbonate shell construction, validating the material type label.
</example>
<example>
ID: B07XYZ123A | Category: CLOTHING
Title: Men's Cotton Blend Casual T-Shirt Blue Medium | Description: [Not provided] | Bullet Points: [Not provided]
Attribute Name: sleeve_length | Generated Label: Long Sleeve
Your Response:
UNKNOWN
Title mentions t-shirt but provides no sleeve length information, and additional product details are unavailable.
</example>
<example>
ID: B09ABC456D | Category: ELECTRONICS
Title: Wireless Bluetooth Headphones with Noise Cancellation
Description: Premium over-ear headphones featuring active noise cancellation technology
Bullet Points: * 30-hour battery life * Bluetooth 5.0 connectivity * Foldable design for portability
Attribute Name: connectivity_technology | Generated Label: USB-C
Your Response:
INCORRECT
Product clearly indicates Bluetooth connectivity, directly contradicting the USB-C label assignment.
</example>
</examples>

<inputs>
<id>{id}</id> <category>{category}</category> <item_name>{title}</item_name>
<description>{description}</description> <bullet_points>{bullet_points}</bullet_points>
<attribute_name>{attribute}</attribute_name> <generated_label>{generated_label}</generated_label>
</inputs>

<instructions>
Evaluate the generated attribute label by:
1. Reviewing all available product information for evidence relating to the specified attribute
2. Determining if the generated label matches, conflicts with, or cannot be verified from the product data
3. Handling incomplete information appropriately when description or bullet points are missing
Classify as:
- CORRECT: Label accurately reflects attribute based on available evidence
- INCORRECT: Label contradicts information in product details
- UNKNOWN: Insufficient data to verify label accuracy
Response format: Respond with exactly one of "CORRECT", "INCORRECT", or "UNKNOWN", followed by a brief explanation.
</instructions>
\end{verbatim}
\end{scriptsize}
\caption{Prompt Version 2: XML-structured format with in-context examples.}
\label{fig:prompt_v2}
\end{figure*}

\subsection{Prompt Version 3: Role-Based with Systematic Guidelines}

\begin{figure*}[t]
\begin{scriptsize}
\begin{verbatim}
## Role
You are an e-Commerce data analyst specializing in product attribute verification and quality assurance.
Your task is to assess whether a generated attribute label correctly represents the specified product attribute
based on available listing information.

## Task Overview
Evaluate the accuracy of a generated attribute label by analyzing product listing data and determining if the label
is supported, contradicted, or cannot be verified by the available information.

## Evaluation Guidelines
Follow these steps systematically:
1. **Review Available Evidence**: Examine all provided product information (title, description, bullet points)
   for any evidence relating to the specified attribute.
2. **Compare Label Against Evidence**: Determine whether the generated label:
   - Matches the product information (explicitly or implicitly)
   - Contradicts the product information
   - Cannot be verified due to insufficient data
3. **Handle Missing Information**: When description or bullet points are unavailable or incomplete,
   acknowledge the limitation in your assessment.

## Classification Categories
- **CORRECT**: The label accurately reflects the attribute based on available evidence in the product listing
- **INCORRECT**: The label contradicts or misrepresents information found in the product details
- **UNKNOWN**: Insufficient data exists to verify the label's accuracy

## Examples
### Example 1: CORRECT
- ID: B08N5WRWNW | Category: LUGGAGE
- Title: <brandname> <makename> Hardside Expandable Luggage with Spinner Wheels, Brushed Anthracite, Carry-On 20-Inch
- Description: Durable polycarbonate shell construction with scratch-resistant texture
- Bullet Points: * 100% Polycarbonate shell * Four multidirectional spinner wheels * TSA-approved combination lock
- Attribute: material_type | Label: Polycarbonate
Assessment: CORRECT - Description and bullet points explicitly confirm polycarbonate shell construction.

### Example 2: UNKNOWN
- ID: B07XYZ123A | Category: CLOTHING
- Title: Men's Cotton Blend Casual T-Shirt Blue Medium | Description: [Not provided] | Bullet Points: [Not provided]
- Attribute: sleeve_length | Label: Long Sleeve
Assessment: UNKNOWN - Title mentions t-shirt but provides no sleeve length information.

### Example 3: INCORRECT
- ID: B09ABC456D | Category: ELECTRONICS
- Title: Wireless Bluetooth Headphones with Noise Cancellation
- Description: Premium over-ear headphones featuring active noise cancellation technology
- Bullet Points: * 30-hour battery life * Bluetooth 5.0 connectivity * Foldable design for portability
- Attribute: connectivity_technology | Label: USB-C
Assessment: INCORRECT - Product clearly indicates Bluetooth connectivity, contradicting USB-C label.

## Product Information to Evaluate
- **ID**: {id} | **Category**: {category} | **Title**: {title}
- **Description**: {description}
- **Bullet Points**: {bullet_points}
- **Attribute Name**: {attribute} | **Generated Label**: {generated_label}

## Response Format
Provide your assessment as: CORRECT/INCORRECT/UNKNOWN followed by a brief explanation (1-2 sentences).
\end{verbatim}
\end{scriptsize}
\caption{Prompt Version 3: Role-based format with systematic guidelines and markdown structure.}
\label{fig:prompt_v3}
\end{figure*}

%% file: sections/appendix_evaluation_settings.tex
\section{Evaluation Settings}
\label{appendix:eval_setting}

\subsection{Judge Qualification Criteria}
\label{appendix:qualification}

Each LLM judge must meet qualification criteria that parallel traditional standards for human auditors. \textit{General competency} establishes baseline capability: models must meet minimum performance thresholds on established benchmarks, such as Global Average $\geq 70\%$ on LiveBench~\citep{white2025livebench}, paralleling education or certification requirements for human auditors. \textit{Task-specific competency} ensures domain relevance: models must demonstrate sufficient understanding of attribute verification through internal benchmark evaluation, analogous to subject matter expertise. \textit{Self-consistency} addresses output reliability: given known variability in LLM outputs due to floating-point precision issues~\citep{yuan2025give}, judges must achieve minimum consistency when generating responses for identical prompts. \textit{Instruction adherence} ensures usable outputs: models must reliably follow specified output formats, paralleling structured reporting requirements for human evaluators.

We stress that these criteria govern the \textit{competence} of individual judges, not their statistical independence. Qualified judges may still fail in correlated ways, and Appendix~\ref{appendix:results_detail} quantifies the extent to which they do.

\subsection{Input Format}

\begin{figure}[t]
\begin{small}
\begin{verbatim}
{
  "id": "X",
  "category": "FILE_FOLDER",
  "text_fields": {
    "title": "<brand>, Lot de 100 pochettes...",
    "features": [
      "Les pochettes A4 en polyéthylène...",
      "Pochettes transparentes..."
    ],
    "description": "Ces chemises en plastique..."
  },
  "attribute_of_interest": {
    "name": "tab.position",
    "value": "Côté"
  }
}
\end{verbatim}
\end{small}
\caption{Example input format for LLM judges. Models receive structured product data with target attribute to verify.}
\label{fig:input_format}
\vskip -0.1in
\end{figure}

Figure~\ref{fig:input_format} shows the input format provided to each judge.

\subsection{Model Configuration}

All models use a low temperature (0.1) to ensure consistent, reproducible outputs. We leave default settings for Top-p and Top-k parameters. 

For OpenAI's GPT OSS reasoning model, we set \newline\texttt{reasoning\_effort="low"} to balance computational cost with performance. This parameter controls the depth of the model's internal chain-of-thought reasoning before producing a final answer. Even at the ``low'' setting, the model demonstrates strong performance on our classification task while significantly reducing inference costs compared to higher reasoning effort levels.

\textbf{Model-Specific Formats.} Each model family requires different API message formats:
\begin{itemize}
    \item \textbf{Claude}: Uses Anthropic's message format with\newline\texttt{anthropic\_version} specification
    \item \textbf{Nova}: Requires nested content structure with explicit text fields
    \item \textbf{OpenAI}: Standard chat completion format with\newline\texttt{reasoning\_effort} parameter
    \item \textbf{Mistral}: Uses instruction-tuned format with \texttt{[INST]} tags
    \item \textbf{DeepSeek/Qwen3/Gemma}: Standard chat message format
\end{itemize}

\subsection{Response Parsing}

Models return responses in different formats requiring specialized parsing:

\begin{itemize}
    \item \textbf{Standard format} (Claude, Nova, Mistral, DeepSeek, Qwen3, Gemma): First line contains the classification label (CORRECT, INCORRECT, or UNKNOWN), followed by reasoning on subsequent lines.
    
    \item \textbf{OpenAI reasoning format}: Response contains\newline \texttt{<reasoning>...</reasoning>} tags encapsulating the model's chain-of-thought, followed by the final classification label. The parser extracts both components separately.
\end{itemize}

Invalid or unparseable responses default to UNKNOWN with the full response content logged for debugging.

\subsection{Evaluation Pipeline}

For each product and judge configuration:

\begin{enumerate}
    \item \textbf{Input formatting}: Product data formatted according to model-specific API requirements
    \item \textbf{API invocation}: Model called via AWS Bedrock with configured parameters
    \item \textbf{Response parsing}: Extract classification and reasoning from model output, handling format variations
    \item \textbf{Validation}: Verify response contains valid classification (CORRECT, INCORRECT, or UNKNOWN); invalid responses default to UNKNOWN
    \item \textbf{Error handling}: API failures and malformed responses are caught and logged, with graceful degradation to UNKNOWN
    \item \textbf{Storage}: Save classification, reasoning, model ID, prompt version, and timestamp
\end{enumerate}

All models were accessed through AWS Bedrock's unified API, ensuring consistent infrastructure across different model providers. The pipeline processes 21 judge configurations (7 models $\times$ 3 prompt variants) per product systematically.

\subsection{Aggregation Details}
\label{appendix:aggregation}

\textbf{Tie resolution.} With 21 judges and three labels, a strict tie is possible (for example a 9/9/3 split) but rare, affecting well under 1\% of samples. Ties are resolved in favour of the label supported by the highest-accuracy judge among those tied, evaluated on the audited subset. Because tied cases are by construction low-agreement, they fall in the stratum we recommend routing to human review, and the choice of tie-breaking rule has a negligible effect on aggregate accuracy.

\textbf{Malformed generator output.} A small fraction of samples emerged from the generation pipeline without a usable three-way label, either because the generator emitted no label, produced an unparseable value, or attached the attribute to a mismatched product identifier. These samples are counted as disagreement cases and routed to expert adjudication, since no synthetic label exists for the arena to agree with. Practitioners reusing this pipeline should expect a comparable residue and budget adjudication capacity for it.

\textbf{Judge-level failures.} Invalid or unparseable judge responses default to UNKNOWN, as described above. This is a conservative choice that pushes affected judges towards abstention rather than towards a substantive label, and it slightly depresses measured per-judge accuracy relative to a design that discarded such responses.

%% file: sections/appendix_cost.tex
\section{Cost Analysis Details}
\label{appendix:costs}

This appendix provides detailed cost breakdowns for the LLM Arena validation framework. All costs are based on AWS Bedrock on-demand pricing as of January 2026.\footnote{\url{https://aws.amazon.com/bedrock/pricing/}}

\subsection{Token Statistics}

Each API call consists of input tokens (the prompt with product data) and output tokens (the model's classification and reasoning). Input token counts vary based on prompt version and product text length; output tokens are estimated at 50 tokens per response (classification label plus brief reasoning).

\begin{table}[h]
\caption{Average input tokens per prompt by language and version}
\label{tab:token_stats}
\begin{center}
\begin{small}
\begin{sc}
\setlength{\tabcolsep}{2.5pt}
\begin{tabular}{llrrrrr}
\toprule
Lang & Ver & Mean & Med. & Min & Max & Std \\
\midrule
DE & V1 & 727 & 664 & 272 & 2554 & 308 \\
 & V2 & 1095 & 1031 & 643 & 2925 & 308 \\
 & V3 & 1229 & 1167 & 776 & 3060 & 309 \\
\midrule
ES & V1 & 688 & 637 & 273 & 2229 & 285 \\
 & V2 & 1056 & 1005 & 639 & 2596 & 285 \\
 & V3 & 1191 & 1140 & 778 & 2733 & 285 \\
\midrule
FR & V1 & 716 & 671 & 274 & 2222 & 291 \\
 & V2 & 1084 & 1039 & 638 & 2587 & 290 \\
 & V3 & 1219 & 1174 & 775 & 2723 & 291 \\
\midrule
IT & V1 & 714 & 665 & 274 & 2079 & 301 \\
 & V2 & 1082 & 1032 & 644 & 2446 & 300 \\
 & V3 & 1217 & 1167 & 780 & 2581 & 300 \\
\bottomrule
\end{tabular}
\end{sc}
\end{small}
\end{center}
\end{table}

Table~\ref{tab:token_stats} presents input token statistics. The three prompt versions differ in complexity: V1 provides basic instructions, V2 adds attribute descriptions, and V3 includes worked examples. This is reflected in token counts, with V3 prompts averaging 70\% more tokens than V1.

Token counts vary across languages due to differences in product text length and linguistic characteristics. German (DE) has the highest average token counts, while Spanish (ES) has the lowest. The standard deviation of approximately 285--308 tokens reflects the diversity in product description lengths across the dataset.

\subsection{Full Arena Costs}

\begin{table}[h]
\caption{Full arena costs by language}
\label{tab:arena_costs_language}
\begin{center}
\begin{small}
\begin{sc}
\setlength{\tabcolsep}{3pt}
\begin{tabular}{lrrrr}
\toprule
Lang. & Products & API Calls & \$/Prod. & Total \\
\midrule
IT & 3,007 & 63,147 & \$0.023 & \$68.80 \\
DE & 3,059 & 64,239 & \$0.023 & \$70.69 \\
ES & 3,159 & 66,339 & \$0.022 & \$70.76 \\
FR & 3,501 & 73,521 & \$0.023 & \$80.25 \\
\midrule
\textbf{All} & \textbf{12,726} & \textbf{267,246} & \textbf{\$0.023} & \textbf{\$290.50} \\
\bottomrule
\end{tabular}
\end{sc}
\end{small}
\end{center}
\end{table}

Table~\ref{tab:arena_costs_language} shows the per-language cost breakdown for running the full 21-judge arena (7 models $\times$ 3 prompt versions). Costs scale linearly with dataset size, with French being most expensive due to having the most products.

\subsection{Scaling Projections}

Table~\ref{tab:scaling_costs} projects arena costs at various scales. At production scale (1M products), the full arena would cost approximately \$22,825—substantially less than equivalent human annotation, which would cost \$100,000--500,000 at typical rates of \$0.10--0.50 per label.

\begin{table}[h]
\caption{Projected arena costs at scale}
\label{tab:scaling_costs}
\begin{center}
\begin{small}
\begin{sc}
\setlength{\tabcolsep}{3.5pt}
\begin{tabular}{rrrr}
\toprule
Products & API Calls & Cost & \$/Prod. \\
\midrule
1K & 21K & \$23 & \$0.023 \\
10K & 210K & \$228 & \$0.023 \\
100K & 2.1M & \$2,283 & \$0.023 \\
1M & 21M & \$22,825 & \$0.023 \\
\bottomrule
\end{tabular}
\end{sc}
\end{small}
\end{center}
\end{table}

\subsection{Cost-Efficient Configurations}

For budget-constrained deployments, reduced ensemble configurations offer significant savings with potential accuracy trade-offs.

\begin{table}[h]
\caption{Alternative ensemble configurations (per 1,000 products)}
\label{tab:reduced_ensembles}
\begin{center}
\begin{small}
\begin{sc}
\setlength{\tabcolsep}{3pt}
\begin{tabular}{lrrr}
\toprule
Configuration & Judges & \$/1K & Savings \\
\midrule
Full Arena (7$\times$3) & 21 & 22.83 & -- \\
Excl. highest-cost & 18 & 11.57 & 49\% \\
Top-3 families (3$\times$3) & 9 & 2.08 & 91\% \\
Single family (3 prompts) & 3 & 0.54--11.26 & 51--97\% \\
Single config & 1 & 0.14--4.39 & 81--99\% \\
\bottomrule
\end{tabular}
\end{sc}
\end{small}
\end{center}
\end{table}

Table~\ref{tab:reduced_ensembles} presents alternative ensemble configurations. Cost varies substantially across model families, with open-weight models typically 10--30$\times$ cheaper than commercial APIs. Using a single model family with all three prompt versions represents 3--49\% of full arena cost depending on family choice.

The trade-off between cost and accuracy presents optimization opportunities. Unanimous agreement cases (36\% of products) could use a smaller initial ensemble, with full arena deployment reserved for uncertain cases. Such adaptive strategies could reduce average cost by 30--50\% while maintaining high accuracy on confident  predictions. However, ablation studies validating accuracy of reduced configurations are required before deployment.

%% file: custom.bib
@misc{negri2025attribute,
  title = {Attribute-Aware Controlled Product Generation with {LLMs} for E-commerce},
  author = {Negri, Virginia and G{\'o}mez, V{\'\i}ctor Mart{\'\i}nez and Balanya, Sergio A and Rajaram, Subburam},
  year = {2025},
  eprint = {2601.04200},
  archivePrefix = {arXiv},
  primaryClass = {cs.CL}
}

@inproceedings{deng2024information,
  author={Deng, Shumin and Ma, Yubo and Zhang, Ningyu and Cao, Yixin and Hooi, Bryan},
  booktitle={2024 IEEE International Conference on Knowledge Graph (ICKG)}, 
  title={Information Extraction in Low-Resource Scenarios: Survey and Perspective}, 
  year={2024},
  volume={},
  number={},
  pages={33-49},
  doi={10.1109/ICKG63256.2024.00013}}

@article{hsu2025leveraging,
  title = {Leveraging Large Language Models for Knowledge-Free Weak Supervision in Clinical Natural Language Processing},
  author = {Hsu, Enshuo and Roberts, Kirk},
  journal = {Scientific Reports},
  volume = {15},
  number = {1},
  pages = {8241},
  year = {2025},
  publisher = {Nature Publishing Group UK London}
}

@misc{sabeh2024exploring,
  title = {Exploring Large Language Models for Product Attribute Value Identification},
  author = {Sabeh, Kassem and Kacimi, Mouna and Gamper, Johann and Litschko, Robert and Plank, Barbara},
  year = {2024},
  eprint = {2409.12695},
  archivePrefix = {arXiv},
  primaryClass = {cs.CL}
}

@inproceedings{satyadharma-etal-2025-auto,
  title = {Auto Prompting Without Training Labels: An {LLM} Cascade for Product Quality Assessment in E-Commerce Catalogs},
  author = {Satyadharma, Soham and Sheikholeslami, Fatemeh and Kaul, Swati and Batur, Aziz Umit and Khan, Suleiman A.},
  editor = {Potdar, Saloni and Rojas-Barahona, Lina and Montella, Sebastien},
  booktitle = {Proceedings of the 2025 Conference on Empirical Methods in Natural Language Processing: Industry Track},
  month = nov,
  year = {2025},
  address = {Suzhou, China},
  publisher = {Association for Computational Linguistics},
  url = {https://aclanthology.org/2025.emnlp-industry.63/},
  doi = {10.18653/v1/2025.emnlp-industry.63},
  pages = {937--953}
}

@inproceedings{mohta2023large,
  title = {Are Large Language Models Good Annotators?},
  author = {Mohta, Jay and Ak, Kenan and Xu, Yan and Shen, Mingwei},
  booktitle = {Proceedings on ``I Can't Believe It's Not Better: Failure Modes in the Age of Foundation Models'' at NeurIPS 2023 Workshops},
  pages = {38--48},
  year = {2023},
  volume = {239},
  series = {Proceedings of Machine Learning Research},
  publisher = {PMLR}
}

@article{de2025fine,
  title = {Fine-Tuned Encoder Models with Data Augmentation Beat {ChatGPT} in Agricultural Named Entity Recognition and Relation Extraction},
  author = {De, Sayan and Sanyal, Debarshi Kumar and Mukherjee, Imon},
  journal = {Expert Systems with Applications},
  volume = {277},
  pages = {127126},
  year = {2025},
  publisher = {Elsevier}
}

@inproceedings{tan2024large,
  title = {Large Language Models for Data Annotation and Synthesis: A Survey},
  author = {Tan, Zhen and Li, Dawei and Wang, Song and Beigi, Alimohammad and Jiang, Bohan and Bhattacharjee, Amrita and Karami, Mansooreh and Li, Jundong and Cheng, Lu and Liu, Huan},
  booktitle = {Proceedings of the 2024 Conference on Empirical Methods in Natural Language Processing},
  month = nov,
  year = {2024},
  address = {Miami, Florida, USA},
  publisher = {Association for Computational Linguistics},
  url = {https://aclanthology.org/2024.emnlp-main.54/},
  doi = {10.18653/v1/2024.emnlp-main.54},
  pages = {930--957}
}

@article{jaradeh2023information,
  title = {Information Extraction Pipelines for Knowledge Graphs},
  author = {Jaradeh, Mohamad Yaser and Singh, Kuldeep and Stocker, Markus and Both, Andreas and Auer, S{\"o}ren},
  journal = {Knowledge and Information Systems},
  volume = {65},
  number = {5},
  pages = {1989--2016},
  year = {2023},
  publisher = {Springer}
}

@article{zhu2024llms,
  title = {{LLMs} for Knowledge Graph Construction and Reasoning: Recent Capabilities and Future Opportunities},
  author = {Zhu, Yuqi and Wang, Xiaohan and Chen, Jing and Qiao, Shuofei and Ou, Yixin and Yao, Yunzhi and Deng, Shumin and Chen, Huajun and Zhang, Ningyu},
  journal = {World Wide Web},
  volume = {27},
  number = {5},
  pages = {58},
  year = {2024},
  publisher = {Springer}
}

@article{xu2024large,
  title = {Large Language Models for Generative Information Extraction: A Survey},
  author = {Xu, Derong and Chen, Wei and Peng, Wenjun and Zhang, Chao and Xu, Tong and Zhao, Xiangyu and Wu, Xian and Zheng, Yefeng and Wang, Yang and Chen, Enhong},
  journal = {Frontiers of Computer Science},
  volume = {18},
  number = {6},
  pages = {186357},
  year = {2024},
  publisher = {Springer}
}

@article{dagdelen2024structured,
  title = {Structured Information Extraction from Scientific Text with Large Language Models},
  author = {Dagdelen, John and Dunn, Alexander and Lee, Sanghoon and Walker, Nicholas and Rosen, Andrew S and Ceder, Gerbrand and Persson, Kristin A and Jain, Anubhav},
  journal = {Nature Communications},
  volume = {15},
  number = {1},
  pages = {1418},
  year = {2024},
  publisher = {Nature Publishing Group UK London}
}

@article{martinez2020information,
  title = {Information Extraction Meets the Semantic Web: A Survey},
  author = {Martinez-Rodriguez, Jose L and Hogan, Aidan and Lopez-Arevalo, Ivan},
  journal = {Semantic Web},
  volume = {11},
  number = {2},
  pages = {255--335},
  year = {2020},
  publisher = {SAGE Publications Sage UK: London, England}
}

@inproceedings{zhang2024automated,
  title = {Automated Mining of Structured Knowledge from Text in the Era of Large Language Models},
  author = {Zhang, Yunyi and Zhong, Ming and Ouyang, Siru and Jiao, Yizhu and Zhou, Sizhe and Ding, Linyi and Han, Jiawei},
  booktitle = {Proceedings of the 30th ACM SIGKDD Conference on Knowledge Discovery and Data Mining},
  pages = {6644--6654},
  year = {2024}
}

@inproceedings{zheng2023judging,
  title = {Judging {LLM}-as-a-Judge with {MT-Bench} and {Chatbot Arena}},
  author = {Zheng, Lianmin and Chiang, Wei-Lin and Sheng, Ying and Zhuang, Siyuan and Wu, Zhanghao and Zhuang, Yonghao and Lin, Zi and Li, Zhuohan and Li, Dacheng and Xing, Eric and others},
  booktitle = {Advances in Neural Information Processing Systems},
  volume = {36},
  year = {2023}
}

@misc{nadas2025synthetic,
  title = {Synthetic Data Generation Using Large Language Models: Advances in Text and Code},
  author = {Nada{\c{s}}, Mihai and Dio{\c{s}}an, Laura and Tomescu, Andreea},
  year = {2025},
  eprint = {2503.14023},
  archivePrefix = {arXiv},
  primaryClass = {cs.CL}
}

@inproceedings{white2025livebench,
  title = {{LiveBench}: A Challenging, Contamination-Limited {LLM} Benchmark},
  author = {White, Colin and Dooley, Samuel and Roberts, Manley and Pal, Arka and Feuer, Ben and Jain, Siddhartha and Shwartz-Ziv, Ravid and Jain, Neel and Saifullah, Khalid and Dey, Siddartha and others},
  booktitle = {Proceedings of the International Conference on Learning Representations},
  year = {2025}
}

@misc{yuan2025give,
  title = {Understanding and Mitigating Numerical Sources of Nondeterminism in {LLM} Inference},
  author = {Yuan, Jiayi and Li, Hongyi and Ding, Xingyu and Xie, Wenting and Li, Yu-Jhe and Zhao, Wenxuan and Wan, Kehan and Shi, Junda and Hu, Xuan and Liu, Ziniu},
  year = {2025},
  eprint = {2506.09501},
  archivePrefix = {arXiv},
  primaryClass = {cs.CL}
}

@techreport{nova_models,
  title = {The {Amazon Nova} Family of Models: Technical Report and Model Card},
  author = {{Amazon Artificial General Intelligence}},
  year = {2024},
  institution = {Amazon},
  url = {https://www.amazon.science/publications/the-amazon-nova-family-of-models-technical-report-and-model-card}
}

@misc{openai2024gpt4technicalreport,
  title = {{GPT}-4 Technical Report},
  author = {OpenAI and Josh Achiam and Steven Adler and Sandhini Agarwal and Lama Ahmad and Ilge Akkaya and Florencia Leoni Aleman and Diogo Almeida and Janko Altenschmidt and Sam Altman and Shyamal Anadkat and others},
  year = {2024},
  eprint = {2303.08774},
  archivePrefix = {arXiv},
  primaryClass = {cs.CL}
}

@misc{TheC3,
  title = {The {Claude} 3 Model Family: {Opus}, {Sonnet}, {Haiku}},
  author = {{Anthropic}},
  year = {2024},
  howpublished = {\url{https://www.anthropic.com/news/claude-3-family}},
  note = {Accessed: 2024-03-14}
}

@misc{seitl2024assessing,
  title = {Assessing the Quality of Information Extraction},
  author = {Seitl, Filip and Kov{\'a}{\v{r}}{\'\i}k, Tom{\'a}{\v{s}} and Mirshahi, Soheyla and Kry{\v{s}}t{\r{u}}fek, Jan and Dujava, Rastislav and Ondrei{\v{c}}ka, Mat{\'u}{\v{s}} and Ullrich, Herbert and Gronat, Petr},
  year = {2024},
  eprint = {2404.04068},
  archivePrefix = {arXiv},
  primaryClass = {cs.CL}
}

@article{Guo_2025,
  title = {{DeepSeek-R1} Incentivizes Reasoning in {LLMs} Through Reinforcement Learning},
  author = {Guo, Daya and Yang, Dejian and Zhang, Haowei and Song, Junxiao and Wang, Peiyi and Zhu, Qihao and Xu, Runxin and Zhang, Ruoyu and Ma, Shirong and Bi, Xiao and others},
  journal = {Nature},
  volume = {645},
  number = {8081},
  pages = {633--638},
  year = {2025},
  publisher = {Springer Science and Business Media LLC},
  doi = {10.1038/s41586-025-09422-z}
}

@misc{gemmateam2025gemma3technicalreport,
  title = {{Gemma} 3 Technical Report},
  author = {{Gemma Team} and Aishwarya Kamath and Johan Ferret and Shreya Pathak and Nino Vieillard and Ramona Merhej and Sarah Perrin and Tatiana Matejovicova and Alexandre Ram{\'e} and Morgane Rivi{\`e}re and others},
  year = {2025},
  eprint = {2503.19786},
  archivePrefix = {arXiv},
  primaryClass = {cs.CL}
}

@misc{mistral_large3_2025,
  title = {Introducing {Mistral} 3: {Mistral Large} 3 and the {Mistral} 3 Model Family},
  author = {{Mistral AI}},
  year = {2025},
  howpublished = {\url{https://mistral.ai/news/mistral-3}},
  note = {Accessed: 2026-02-04}
}

@inproceedings{yang2021maveproductdatasetmultisource,
  title = {{MAVE}: A Product Dataset for Multi-Source Attribute Value Extraction},
  author = {Yang, Li and Wang, Qifan and Yu, Zac and Kulkarni, Anand and Sanghai, Sumit and Shu, Bin and Elsas, Jon and Kanagal, Bhargav},
  booktitle = {Proceedings of the Fifteenth ACM International Conference on Web Search and Data Mining},
  pages = {1256--1265},
  year = {2022},
  publisher = {Association for Computing Machinery},
  address = {New York, NY, USA},
  doi = {10.1145/3488560.3498377},
  series = {WSDM '22}
}

@misc{bai2025qwen3vltechnicalreport,
  title = {{Qwen3-VL} Technical Report},
  author = {Shuai Bai and Yuxuan Cai and Ruizhe Chen and Keqin Chen and Xionghui Chen and Zesen Cheng and Lianghao Deng and Wei Ding and Chang Gao and Chunjiang Ge and others},
  year = {2025},
  eprint = {2511.21631},
  archivePrefix = {arXiv},
  primaryClass = {cs.CV}
}

@article{landis1977measurement,
  title = {The Measurement of Observer Agreement for Categorical Data},
  author = {Landis, J. Richard and Koch, Gary G.},
  journal = {Biometrics},
  pages = {159--174},
  year = {1977},
  publisher = {JSTOR}
}

@misc{verga2024replacing,
  title = {Replacing Judges with Juries: Evaluating {LLM} Generations with a Panel of Diverse Models},
  author = {Verga, Pat and Hofstatter, Sebastian and Althammer, Sophia and Su, Yixuan and Piktus, Aleksandra and Arkhangorodsky, Arkady and Xu, Minjie and White, Naomi and Lewis, Patrick},
  year = {2024},
  eprint = {2404.18796},
  archivePrefix = {arXiv},
  primaryClass = {cs.CL},
  url = {https://arxiv.org/abs/2404.18796}
}

@misc{kohli2026nine,
  title = {Nine Judges, Two Effective Votes: Correlated Errors Undermine {LLM} Evaluation Panels},
  author = {Kohli, G.},
  year = {2026},
  eprint = {2605.29800},
  archivePrefix = {arXiv},
  primaryClass = {cs.CL},
  url = {https://arxiv.org/abs/2605.29800}
}

@article{northcutt2021pervasive,
  title = {Pervasive Label Errors in Test Sets Destabilize Machine Learning Benchmarks},
  author = {Northcutt, Curtis G. and Jiang, Lu and Chuang, Isaac L.},
  journal = {Journal of Artificial Intelligence Research},
  volume = {71},
  pages = {565--619},
  year = {2021}
}

@inproceedings{zhang-etal-2025-leveraging-product,
  title = {Leveraging Product Catalog Patterns for Multilingual {E}-commerce Product Attribute Prediction},
  author = {Zhang, Bryan and Khan, Suleiman and Walter, Stephan},
  editor = {Potdar, Saloni and Rojas-Barahona, Lina and Montella, Sebastien},
  booktitle = {Proceedings of the 2025 Conference on Empirical Methods in Natural Language Processing: Industry Track},
  month = nov,
  year = {2025},
  address = {Suzhou (China)},
  publisher = {Association for Computational Linguistics},
  pages = {267--275}
}
